%% file: main.tex
\documentclass[journal]{IEEEtran}
\IEEEoverridecommandlockouts

\usepackage{amssymb}
\usepackage{float}
\usepackage{makeidx}
\usepackage{amsmath,amsfonts}
\usepackage[tight,footnotesize]{subfigure}
\usepackage{textcomp}
\usepackage{bbm}
\usepackage{epsfig}
\usepackage{epsf}
\usepackage{psfrag}
\usepackage{verbatim}
\usepackage{color}
\usepackage{multirow}
\usepackage{tabularx}
\usepackage{booktabs}
\usepackage{array}
\usepackage{soul}
\usepackage{footnote}
\usepackage{cite}
\usepackage{dblfloatfix}

\usepackage{xcolor}
\usepackage{color, colortbl}
\usepackage[colorlinks,bookmarksnumbered,citecolor=orange,urlcolor=orange]{hyperref}
\usepackage{graphicx}
\graphicspath{{Figures}}
\DeclareGraphicsExtensions{.pdf,.png}
\usepackage{bigstrut}
\usepackage[english]{babel}

\usepackage{csquotes}

\usepackage{algorithm, setspace}
\usepackage{amsthm}
\usepackage{algpseudocode}
\usepackage{url}
\usepackage{multirow}
\usepackage{float}
 \hypersetup{
     colorlinks=true,
     linkcolor=orange,
     filecolor=orange,
     citecolor=orange,      
     urlcolor=orange,
     }
\usepackage[T1]{fontenc}
\usepackage{balance}

\newcommand{\p}[1]{\smallskip \noindent \textbf{{#1}.}}

\newcommand{\fig}[1]{Figure~\ref{fig:#1}}

\usepackage{tcolorbox}

\title{\LARGE

Wrapping Haptic Displays Around Robot Arms to Communicate Learning

}

\author{
Antonio Alvarez Valdivia$^{1}$, Soheil Habibian$^{2}$, Carly A. Mendenhall$^{1}$, Francesco Fuentes$^{1}$, Ritish Shailly$^{2}$, \\ Dylan P. Losey$^{2}$, and Laura H. Blumenschein$^{1}$
\thanks{This work is supported in part by NSF Grants $\#2129201$ and $\#2129155$ and by the NSF Graduate Research Fellowship Program.}
\thanks{$^{1}$Mechanical Engineering, Purdue University, Lafayette, IN 47901. 
        {\texttt{\{alvar168, cmenden, ffuente, lhblumen\}@purdue.edu}}}%
\thanks{$^{2}$These authors are members of the Collaborative Robotics Lab (\href{https://collab.me.vt.edu/}{Collab}), Dept. of Mechanical Engineering, Virginia Tech, Blacksburg, VA 24061.
\newline
{\texttt{\{habibian, rshailly, losey\}@vt.edu}}}

}

\begin{document}
\maketitle


\begin{abstract}
Humans can leverage physical interaction to teach robot arms. As the human kinesthetically guides the robot through demonstrations, the robot learns the desired task. While prior works focus on how the robot learns, it is equally important for the human teacher to understand what their robot is learning. Visual displays can communicate this information; however, we hypothesize that visual feedback alone misses out on the \textit{physical connection} between the human and robot. In this paper we introduce a novel class of \textit{soft haptic displays} that wrap around the robot arm, adding signals without affecting that interaction. We first design a pneumatic actuation array that remains flexible in mounting. We then develop single and multi-dimensional versions of this wrapped haptic display, and explore human perception of the rendered signals during psychophysic tests and robot learning. We ultimately find that people \textit{accurately distinguish} single-dimensional feedback with a Weber fraction of $11.4\%$, and identify multi-dimensional feedback with $94.5\%$ accuracy. When physically teaching robot arms, humans leverage the single- and multi-dimensional feedback to provide better demonstrations than with visual feedback: our wrapped haptic display decreases teaching time while increasing demonstration quality. This improvement depends on the location and distribution of the wrapped haptic display. 

\end{abstract}


\begin{IEEEkeywords}
Haptic Display, Learning from Demonstration, Tactile Devices
\end{IEEEkeywords}


\input{intro}
\input{related}
\input{design}
\input{study1-purdue}
\input{study2-vt}
\input{study3-purdue}

\input{study4-vt}
\input{conclusion}


\bibliographystyle{IEEEtran}
\bibliography{references}


\vspace{-1.5cm}
\begin{IEEEbiography}[{\includegraphics[width=1in,height=1.25in,clip,keepaspectratio]{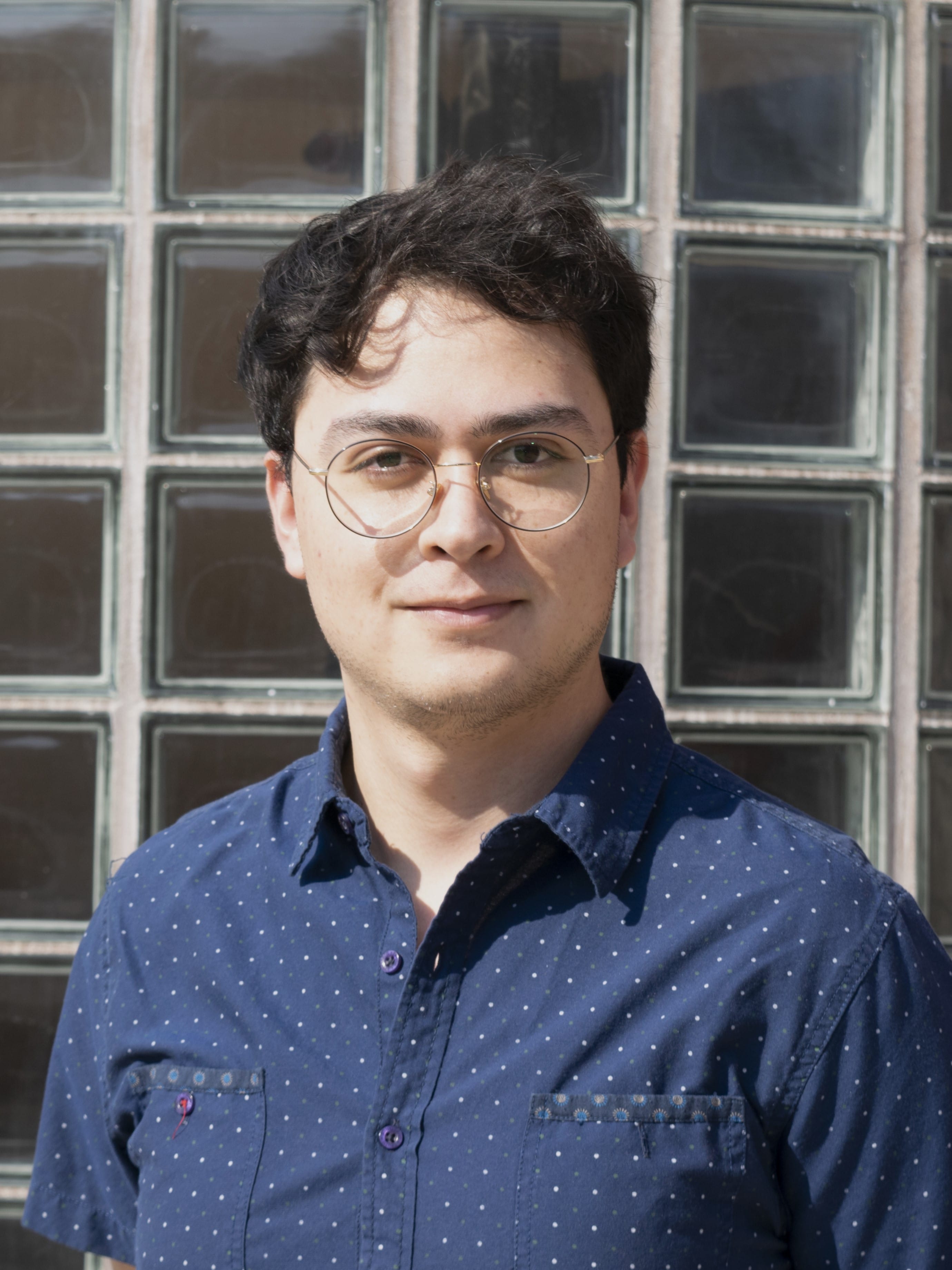}}]{Antonio Alvarez Valdivia}
(Student Member, IEEE) is a Ph.D. student in Mechanical Engineering at Purdue University. He received the B.S. degree in Mechanical Engineering at Iowa State University, Ames, IA, USA, in 2021. He is currently a Research Assistant in the RAAD Lab, Department of Mechanical Engineering, Purdue University. His interests include soft haptics, soft robotics, and mechatronics.
\end{IEEEbiography}
\vspace{-1cm}

\begin{IEEEbiography}[{\includegraphics[width=1in,height=1.25in,clip,keepaspectratio]{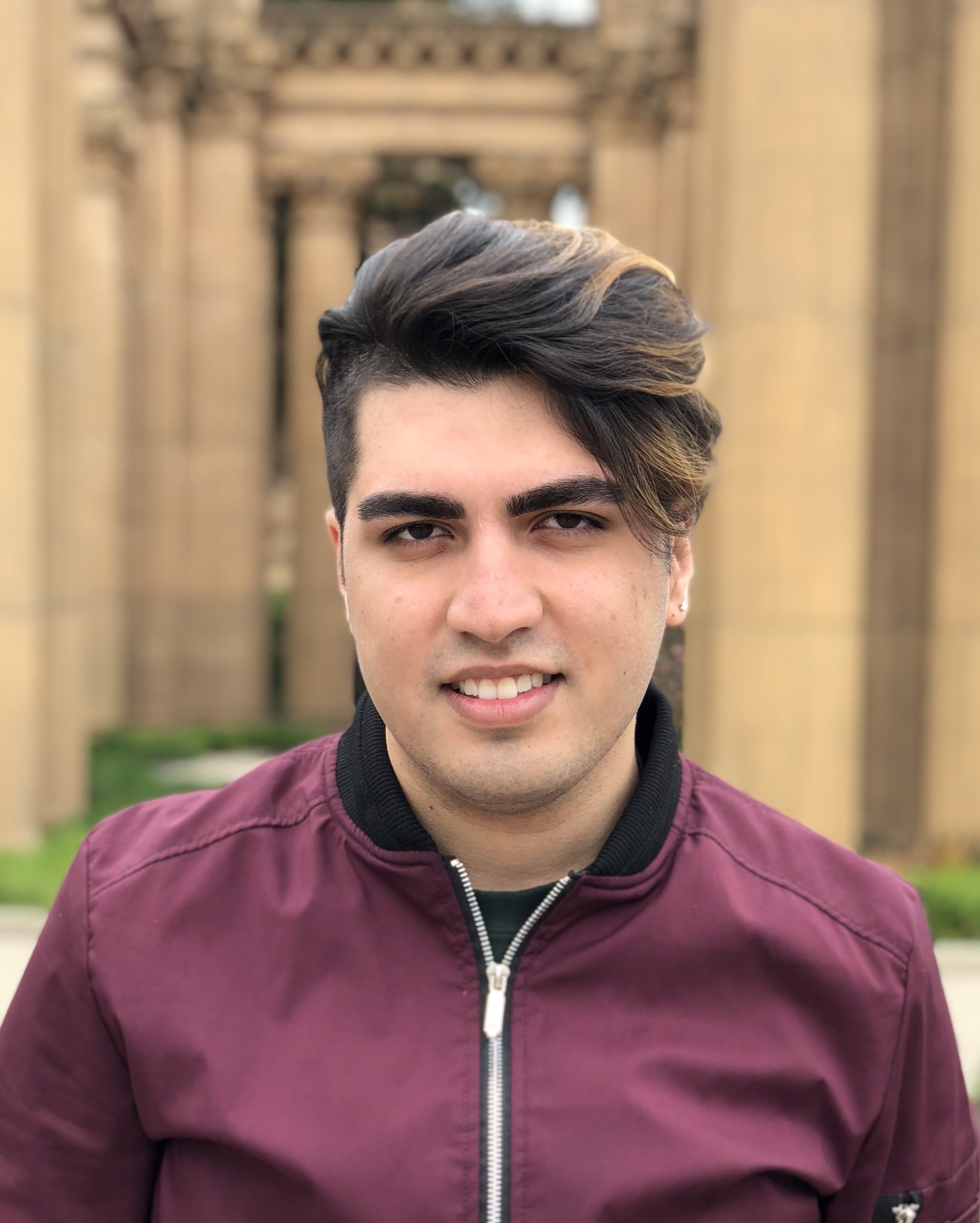}}]{Soheil Habibian}
(Student Member, IEEE) is a Ph.D. student in mechanical engineering at Virginia Tech, VA. He received the B.Sc. degree from Qazvin Azad University, Iran in $2015$, and M.Sc. degree from Bucknell University, PA in $2020$, both in mechanical engineering. He is currently a Research Assistant with Collaborative Robotics Laboratory, Department of Mechanical Engineering, Virginia Tech. His research interests include robot learning, human-robot teams, and artificial intelligence.
\end{IEEEbiography}
\vspace{-1cm}

\begin{IEEEbiography}[{\includegraphics[width=1in,height=1.25in,clip,keepaspectratio]{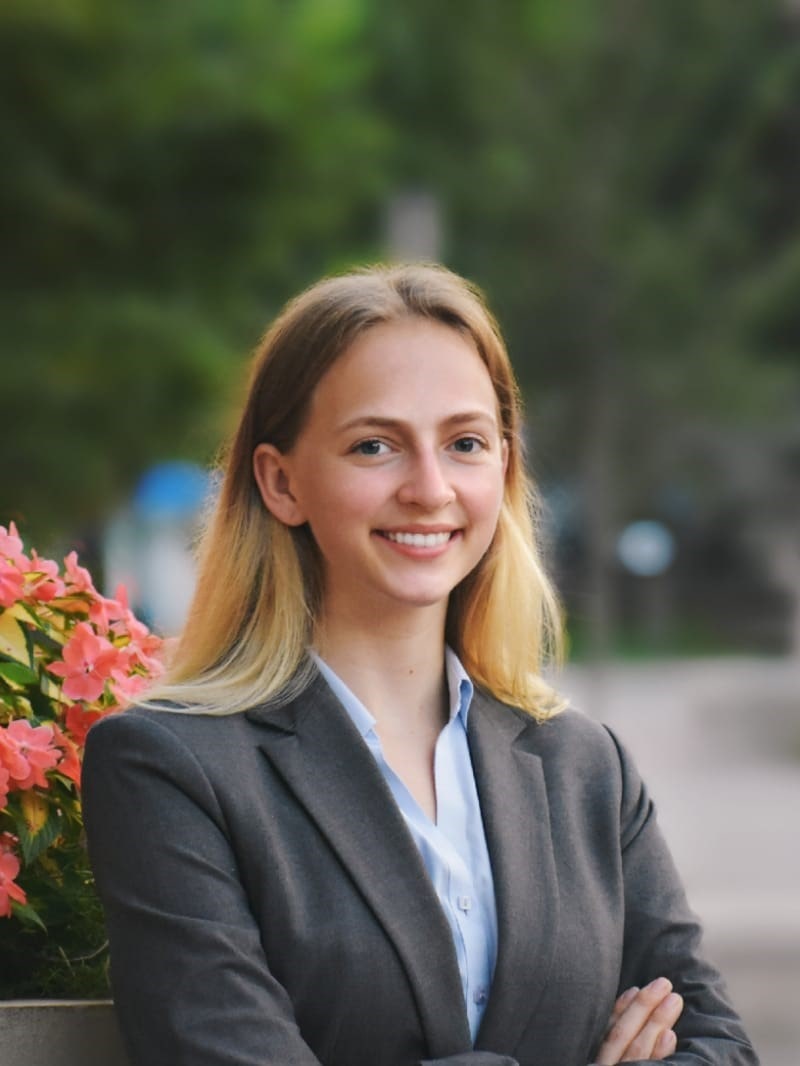}}]{Carly A. Mendenhall} is a Ph.D. student in Mechanical Engineering at Purdue University. She received the B.S. degree in Mechanical Engineering from Purdue University, West Lafayette, IN, USA, in 2022. Since 2021, she has been working jointly in the RAAD Lab and the Tepole Mechanics and Mechanobiology Lab at Purdue University, where she is currently a Research Assistant. Her research interests include soft haptics, soft robotics, and computational modeling of soft materials.
\end{IEEEbiography}
\vspace{-1cm}

\begin{IEEEbiography}[{\includegraphics[width=1in,height=1.25in,clip,keepaspectratio]{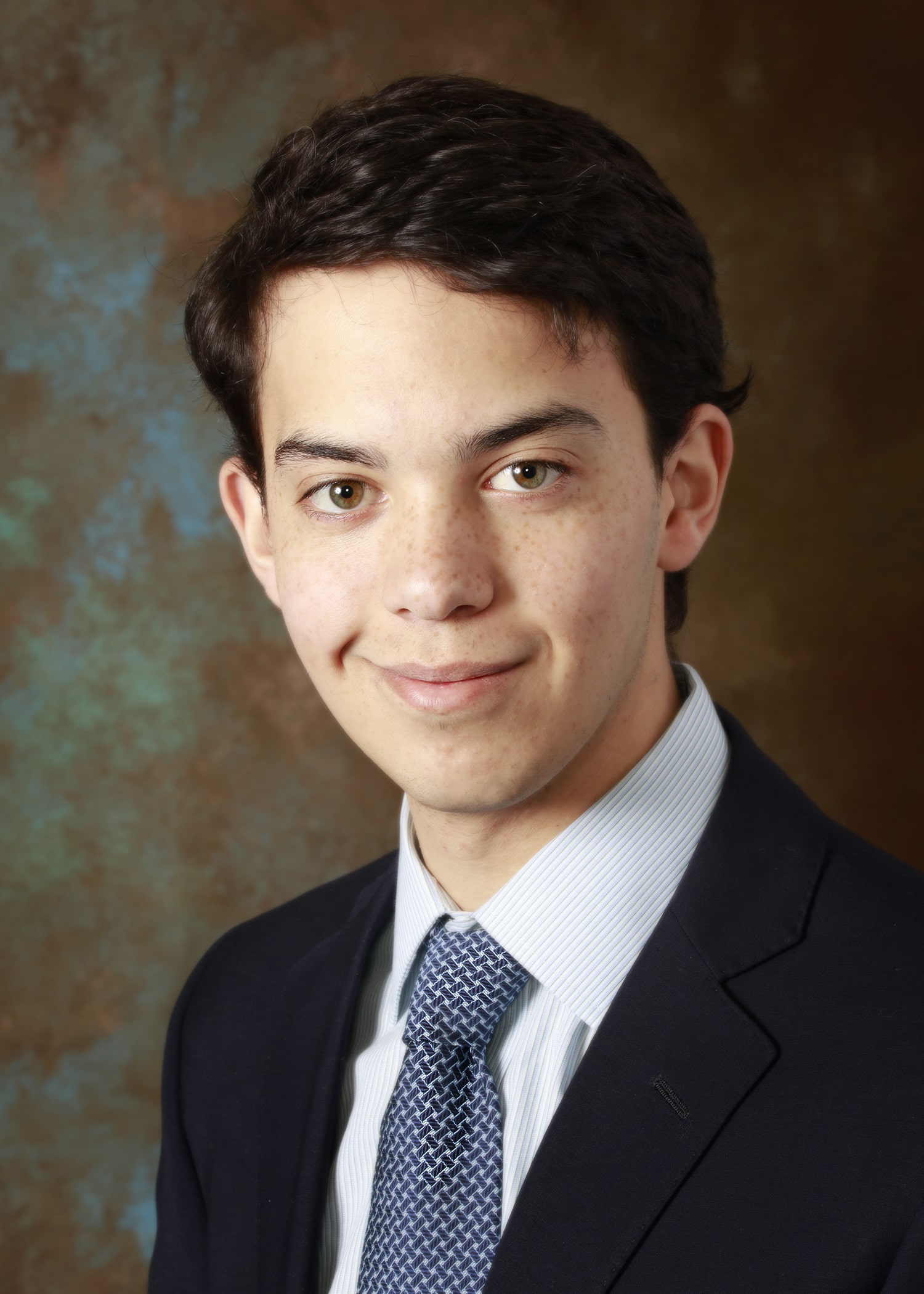}}]{Francesco Fuentes} received his B.S. degree in Mechanical Engineering from Marquette University, Milwaukee, WI, in 2020. Since then, he has been pursuing a Ph.D. in Mechanical Engineering at Purdue University at the RAAD Lab. His research interests include soft robotics, bio-inspired robots, and growing robots.
\end{IEEEbiography}
\vspace{-1cm}

\begin{IEEEbiography}[{\includegraphics[width=1in,height=1.25in,clip,keepaspectratio]{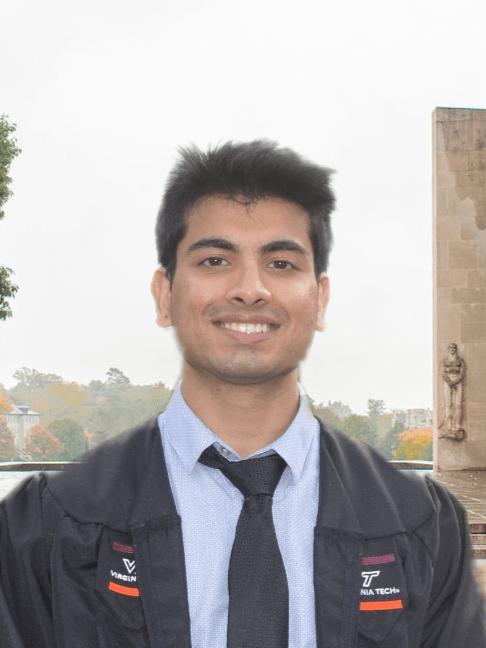}}]{Ritish Shailly}
(Student Member, IEEE) received the B.S. degree in Mechanical Engineering from Virginia Polytechnic Institute and State University (Virginia Tech), Blacksburg, VA, USA in 2021. He worked in Collaborative Robotics Lab, Virginia Tech from 2021 to 2022 and is currently a M.S. student in Mechanical Engineering at Virginia Tech. His research interests include haptics, robot learning, and human perception.
\end{IEEEbiography}
\vspace{-1cm}

\begin{IEEEbiography}[{\includegraphics[width=1in,height=1.25in,clip,keepaspectratio]{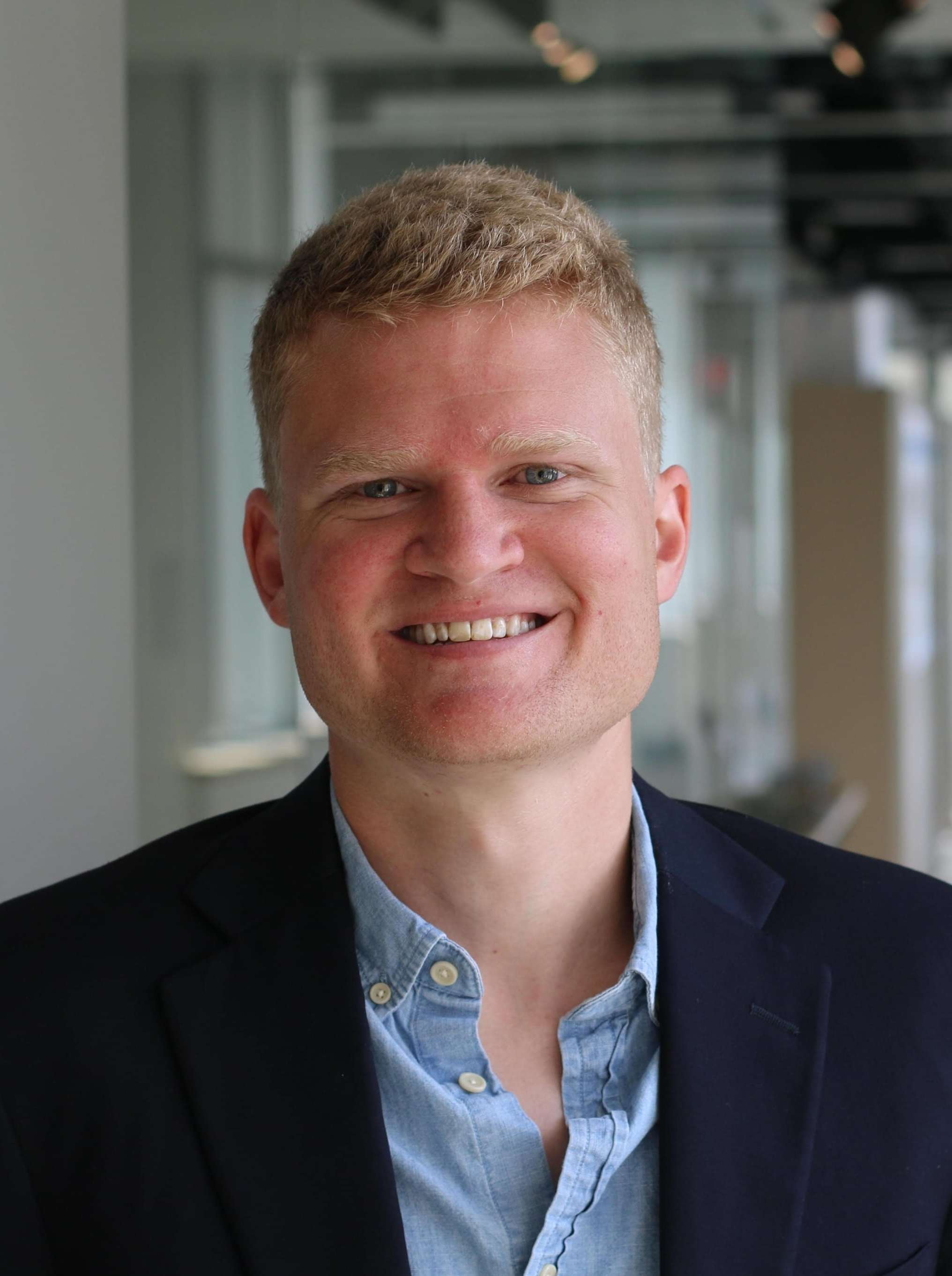}}]{Dylan P. Losey} is an Assistant Professor in the Mechanical Engineering Department at Virginia Tech. His research group develops learning and control algorithms for robots that interact with people. 

Dylan was previously a postdoctoral scholar at Stanford University. He received his doctoral degree from Rice University in 2018, and his bachelor’s degree from Vanderbilt University in 2014. He was awarded the 2020 Conference on Robot Learning Best Paper Award and the 2017 IEEE/ASME Transactions on Mechatronics Best Paper Award.
\end{IEEEbiography}
\vspace{-1cm}

\begin{IEEEbiography}[{\includegraphics[width=1in,height=1.25in,clip,keepaspectratio]{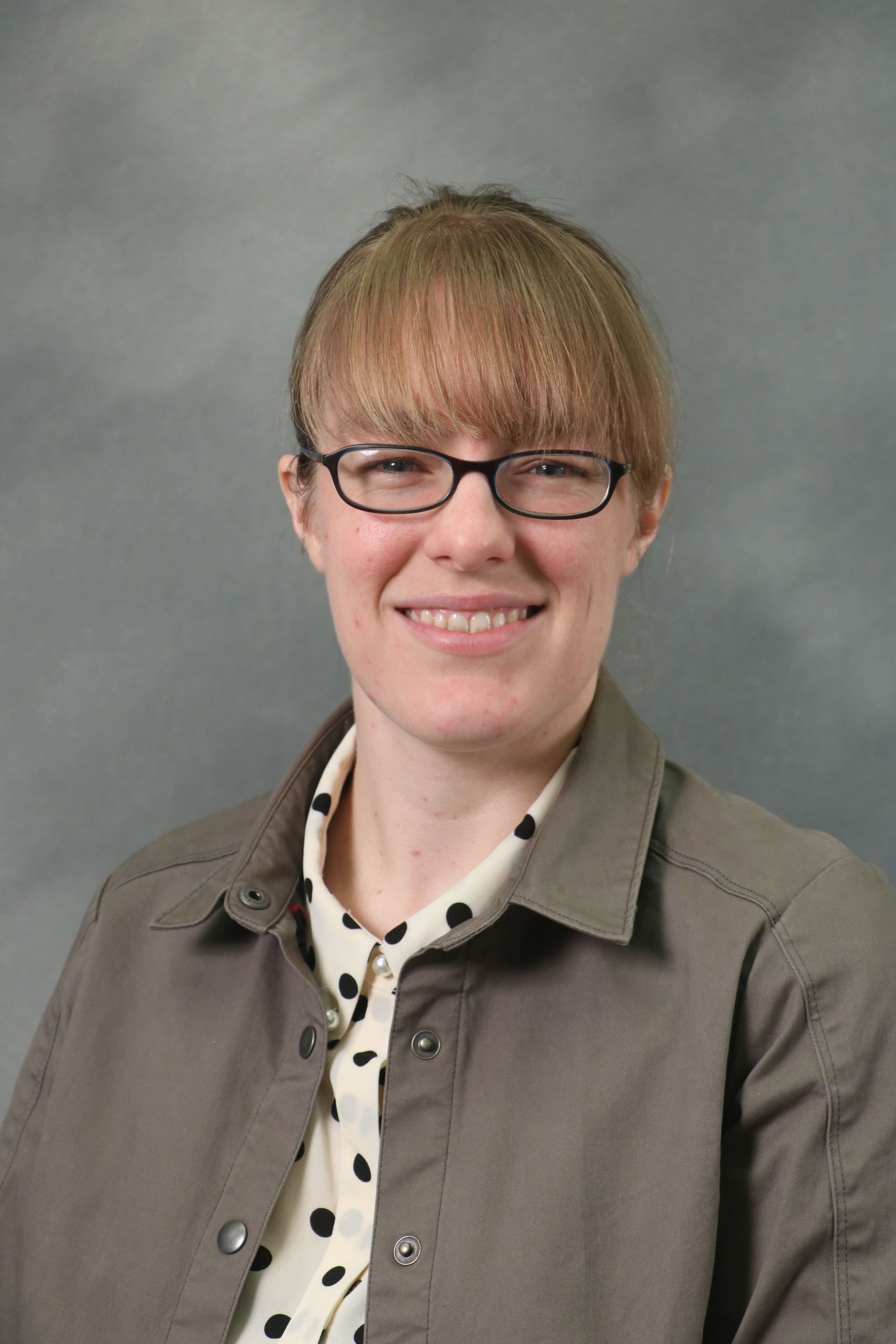}}]{Laura H. Blumenschein}
(Member, IEEE) received the B.S. and M.S. degrees from Rice University, Houston, TX, USA in 2015 and 2016, respectively, and the Ph.D. degree from Stanford University, Stanford, CA, USA in 2019 under the supervision of Prof. A. Okamura, all in mechanical engineering.

She is an Assistant Professor of Mechanical Engineering at Purdue University. Her research interests include soft robotics, actuator design, modeling, haptics, and growing robots.
\end{IEEEbiography}

\end{document}

%% file: intro.tex
\section{Introduction}

\begin{figure}[t]
	\begin{center}
		\includegraphics[width=0.95\columnwidth]{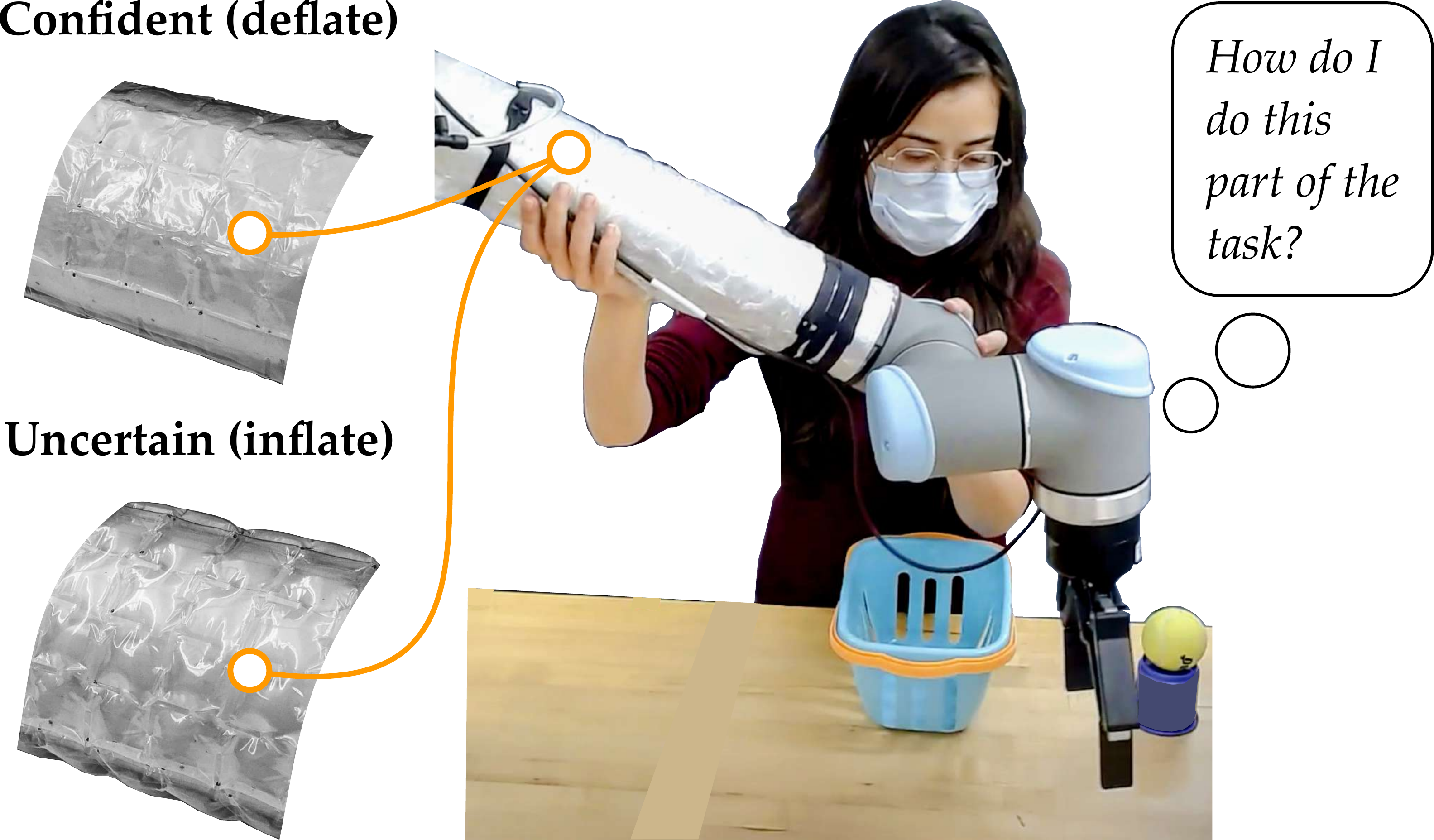}
		\vspace{-0.5em}
		\caption{Human physically teaching a robot arm. We wrap a soft pneumatic display around the arm and render haptic signals by controlling the pressure of the display. The robot learner leverages this haptic display in real-time to communicate the parts of the task that it is confident about, as well as the parts where it is uncertain and needs additional guidance.}
		\label{fig:front}
	\end{center}
 	\vspace{-2.0em}
\end{figure}

Imagine teaching a rigid robot arm to clean objects off a table (see \fig{front}). One intuitive way for you to teach this robot is through \textit{physical interaction}: you push, pull, and guide the arm along each part of the task. Of course, the robot may not learn everything from a single demonstration, and so you show multiple examples of closing shelves, removing trash, and sorting objects. As you kinesthetically teach the robot you are faced with two questions: i) has the robot learned enough to clear the table by itself and ii) if not, what features of the task is the robot still uncertain about?

While existing work enables robots to learn from physical human interaction \cite{argall2009survey,akgun2012keyframe,pastor2009learning,losey2021physical}, having the robot effectively provide \textit{real-time feedback} to human teachers remains an open problem. Ideally, this feedback should not be cumbersome or distracting (i.e., the human must be able to focus on guiding the robot) and should be easily interpretable (i.e., the human must be able to clearly distinguish between signals). These requirements present a tradeoff for haptic feedback as human fingertips provide the densest mechanoreceptors, but placing rigid devices at the hand will impact task performance. Recent research has created communication channels by instead wrapping \textit{haptic devices} around the human's arm \cite{che2020efficient, mullen2021communicating, dunkelberger2020multisensory}, but locating feedback at unrelated locations on the human's body can create a disconnect with the task. 

Our insight is that --- instead of asking the human teacher to wear a feedback device or watch a computer monitor ---
\begin{center}\vspace{-0.3em}
\textit{We can take advantage of the preexisting physical contact between the human and robot through slim form-factor soft haptic displays that \emph{wrap} around the robot arm.}\vspace{-0.3em}
\end{center}
Accordingly, in this paper we develop and analyze wrapped haptic displays for communicating robot learning based on soft robotic principles. We distribute these soft displays along rigid robot arms so that the human can physically interact with the robot to demonstrate a task while simultaneously perceiving the robot's feedback. We actively control the \textit{pressures} of the pneumatic display to render where in the task and what features of the task the robot is \textit{uncertain} about: the display inflates for regions and features of the task where the robot is unsure about its actions (and needs additional human teaching), and deflates where the robot is confident about the task (and does not need any additional human guidance). Our hypothesis is that --- because the soft wrapped display creates a channel for communication on any surface without impacting the task --- humans will be able to more intuitively and robustly use this feedback with a greater level of focus compared to other feedback modalities. We experimentally demonstrate that this pressure-based feedback enables humans i) to determine whether the robot has learned enough to be deployed and ii) to identify parts of the task where kinesthetic teaching is still required. Additionally, we demonstrate the importance of the location and distribution of the feedback on the robot arm for creating this improvement. \textcolor{black}{An interface that provides such intuitive and robust feedback about a robot’s latent state in real time could potentially be introduced to factory settings that feature learning systems. Such displays would allow everyday workers with no programming or robot-related training to interact with learning systems.}

Parts of this work were previously published in \cite{valdivia2022wrapped}, which presented the experimental results for our one degree-of-freedom (DoF) haptic display. This current paper builds on that initial research by demonstrating the design, analysis, and application of \textit{multi-DoF} spatial signals localized or distributed along the robot arm, as well as a follow-up analysis of the $1$-DoF device. Overall, we make the following contributions:

\p{Developing Wrapped Haptic Display} We design and build a compliant pneumatic haptic device that wraps around and conforms to the robot, providing haptic stimuli that are localized to the robot arm and distributed along its geometry. This device is manufactured using soft, flexible pouches that render haptic signals through pressure.

\p{Measuring User Ability to Perceive Wrapped Displays} We perform a psychophysics study to find the range of pressures that humans can distinguish. We report the just noticeable difference (JND) for pressures rendered by the soft display.

\p{Applying Wrapped Displays to Communicate Learning} We ask participants to kinesthetically teach a robot arm while the robot provides real-time feedback about its learning. We map the robot's uncertainty to the pressure of our wrapped display. Compared to a graphical user interface, wrapped haptic display feedback leads to faster and more informative human teaching, and is subjectively preferred.

\p{Extension on Wrapped Displays to Multiple Degrees of Freedom} We generalize the wrapped display design to create multi-degree of freedom displays. These displays can be configured to fit different robotic manipulator geometries and to change the interconnections between pouches.

\p{Measuring Effect of Display Distribution on User Perception} We perform a psychophysics study to understand how the spatial distribution of the wrapped haptic display signals affects the accuracy and speed of signal identification. We demonstrate a tradeoff between speed of identification and accuracy as signals are spread further apart.

\p{Measuring Effect of Display Distribution of Multi-Degree of Freedom Displays for Communicating Learning} We repeated the kinesthetic teaching task with three degree of freedom displays, confirming that users still improve demonstrations over baseline as signal complexity increases. When comparing different options to distribute feedback in 3-DoF displays, users performed better with and subjectively preferred wrapped display layouts where all feedback was displayed the small area where contact was already occurring instead of distributed in larger areas along the robot arm.

%% file: related.tex
\section{Related Work}

In this paper we introduce a wrapped haptic display for communicating robot learning in real-time during physical human-robot interaction. We build on previous research for kinesthetic teaching, haptic interfaces, and soft displays.

\p{Kinesthetic Teaching} Humans can show robot arms how to perform new tasks by physically demonstrating those tasks on the robot \cite{argall2009survey,akgun2012keyframe,pastor2009learning,losey2021physical}. As the human backdrives it, the robot records the states that it visits and the human's demonstrated actions at those states. The robot then learns to imitate the human's actions and perform the task by itself \cite{ross2011reduction}. One important output of the learning process is the robot’s \textit{uncertainty} about the task. The uncertainty can be measured as the robot’s \textit{overall} confidence in \textit{what} to do at different states \cite{hoque2021thriftydagger,menda2019ensembledagger}, or also measure the robot’s confidence on \textit{how} to perform the task \cite{hough2017s, cakmak2012designing, basu2018learning, habibian2022here}. In this paper we explore how robots should \textit{communicate} their learning uncertainty back to the human teacher. Keeping the human up-to-date with what the robot has learned builds trust and improves teaching \cite{hellstrom2018understandable}. Outside of physical human-robot interaction, prior research has developed multiple non-haptic modalities to communicate robot learning and intent: these include robot motion \cite{dragan2013legibility}, graphical user interfaces \cite{huang2019enabling}, projections into the environment \cite{andersen2016projecting}, and augmented reality headsets \cite{walker2018communicating}. Within a teleoperation domain, our recent work suggests that \textit{haptic interfaces} are particularly effective at communicating low-dimensional representations of robot learning \cite{mullen2021communicating}. Here we will leverage these results to develop a real-time feedback interface \textit{specifically for} kinesthetic teaching.

\p{Haptics for Information Transfer}
When using haptics to communicating features of robot learning, the type of information being transferred is important to consider. While haptic devices have a general goal of stimulating the human sense of touch, haptics has also been applied to communicate \textit{robot intent} or similar social features. For instance --- when studying how humans and robots should interact in shared spaces --- prior works have used haptics to explicitly convey the robot's intended motion or actions \cite{che2020efficient, cini2021relevance, casalino2018operator, grushko2021improved}.  Recent work has shown that, given appropriate context, complex human-to-human social touch signals, like stroking \cite{nunez2020investigating, muthukumarana2020touch}, hugging \cite{HuggyPajamaTeh}, dancing \cite{kobayashi2022whole}, and emotional communication \cite{SalvatoTOH2021, ju2021haptic, rognon2022linking}, can be replicated and understood in a wearable format. \textcolor{black}{Some other work has shown the use of haptic interfaces for high information tasks, like assisting navigation through rendering patterns with a certain meaning \cite{paneels2013WHC} or using haptic signals to reduce the distraction from visual displays in human-robot collaborative task scheduling \cite{maderna2022flexible}.} Lastly, work has shown communicating alerts with different urgency levels in car driving \cite{di2020purring, locken2015tacticar} and communicating contact events in teleoperation and AR/VR through hand-held haptic devices \cite{choi2018claw, mintchev2019portable}. These past works suggest that a wide range of social and collaborative information can be transferred using haptics with appropriate design of the interface and signals.

\p{Soft Haptic Devices} 
Soft haptic devices offer an attractive option for human-robot communication due to their compliance and adaptability, through the \textit{flexibility} of the interface or the \textit{compliance} of the actuators. A range of compliant actuation types have been use for haptic devices: pneumatic actuation \cite{raitor2017wrap,HapWRAP2018, do2021macro}, shape memory alloys \cite{muthukumarana2020touch}, dielectric elastomers \cite{zhao2020wearable}, and fluidic elastomers \cite{barreiros2018fluidic}. 
Soft wearable fingertip devices have targeted a range of stimuli in the skin \cite{yin2021wearable}, such as vibrations \cite{ji2021untethered, feng2017submerged, hinchet2018dextres}, indentation \cite{leonardis2022parallel, boys2018soft, hwang2017design}, skin-stretch \cite{minamizawa2010simplified, leonardis2015wearable}, or combinations of those \cite{zhakypov2022fingerprint, leroy2020multimode, youn2021wearable, giraud2022haptigami}. Soft haptic approaches scale easily to increased areas of stimulation; haptic surfaces using arrays of actuators and sensors show scaling to fit varied areas. These developments have typically used rigid elements, such as NFC electronics \cite{yu2019skin}, thin-film strain sensors \cite{sonar2020closed}, and piezo films \cite{suh2014soft}, embedded in cloth and silicone layers to create bi-directional interfaces. These rigid elements can limit the flexibility of the device, and lead to issues with wear over time and comfort. Some tabletop haptic displays have used pneumatically actuated soft composite materials \cite{yao2013pneui} or pneumatic actuation with particle jamming \cite{stanley2015controllable} to control the shape and mechanical properties of surfaces, leading to complex signals and comfortable interaction.

Soft haptic interfaces also support a range of device types distinguished by method of interaction: graspable, wearable, or touchable \cite{culbertson2018haptics}. This method can have a large impact on the usability of the devices. 
Fingertip worn devices provide high fidelity and interpretable signals \cite{yin2021wearable,hinchet2018dextres,zhakypov2022fingerprint}. These devices are popular for virtual reality where physical contact with the real world is unlikely; in other applications they can reduce the user's ability to use there hands. This motivates wearable devices for other body areas, such as hand dorsal \cite{chossat2019soft, wang2019three}, wrists/forearms \cite{raitor2017wrap, muthukumarana2020touch, HapWRAP2018}, or gloves that cover the whole hand \cite{in2015exo}. \textcolor{black}{Our recent work has demonstrated the use of inflatable pouches to create wearable haptic interfaces that provide feedback to humans in the form of distributed spatial inflation \cite{do2021macro}.} Placing haptic signals \textit{directly on the human body} enables the human to move about the space while receiving real-time feedback; but, as feedback is moved away from the fingertip and physically separated from the task, it potentially requires additional mental energy to decode the intended message. A different approach has focused on developing touchable haptic surfaces consisting of arrays of actuators and sensors \cite{yu2019skin, sonar2020closed, suh2014soft}. These devices use the fingertip mechanoreceptors without burdening the user's hands. Soft touchable displays allow installation of haptic interfaces in common touch areas, like car steering wheels \cite{di2020purring}. While not a haptic display, recent work showed pneumatic actuators wrapped around robot arms to visualize the weight load carried by the robot \cite{larsen2022Roso}. Based on this past work, we target a touchable device placed at the point of human-robot interaction, and use soft pneumatic actuation to maximize the flexibility and transparency of the display.

%% file: design.tex
\section{Developing a Wrapped Haptic Display} \label{Haptic Display and Design}



We first aim to design a soft haptic display that can wrap around a robot arm, conforming to the surface and adding a haptic interface to existing points of contact between the human and robot. This section describes the identification of three critical requirements (low volume, fast inflation, and textured surface). With these requirements, we outline two designs built on the same underlying principle: a 1-DoF display with a large contact area and a N-DoF design with multiple, reduced-width, "ring" sleeves. Finally, we describe the how these wrapped haptics displays were implementated. 

\begin{figure*}[t]
	\begin{center}
		\includegraphics[width=1.8\columnwidth]{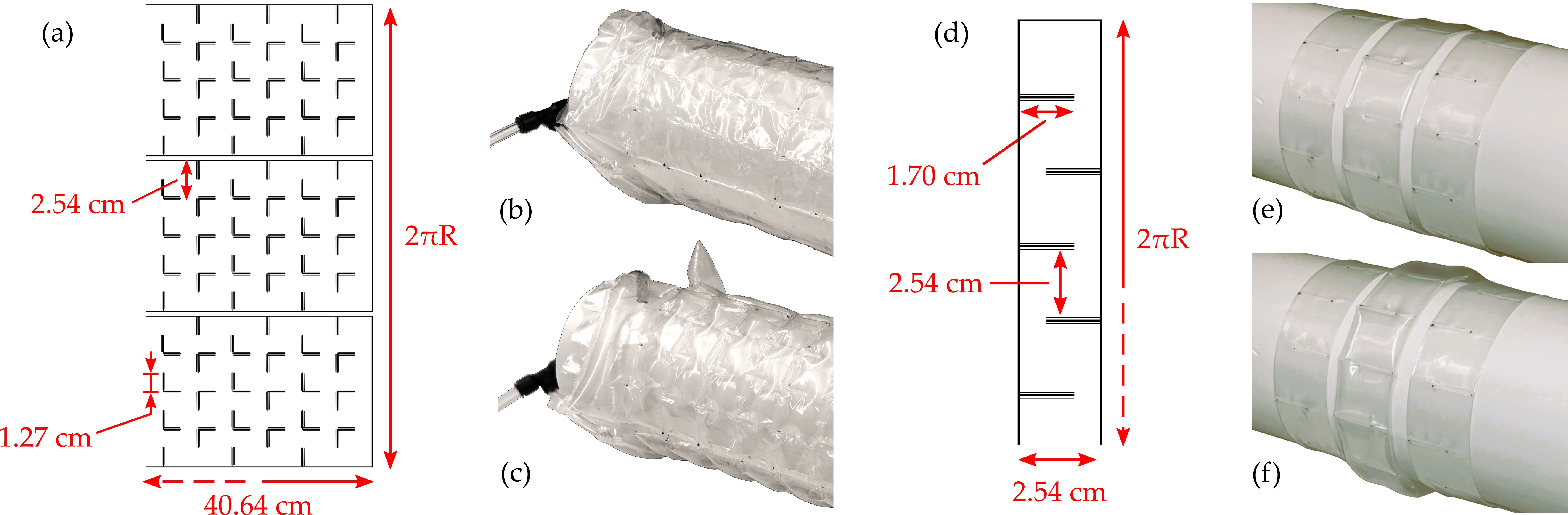}
		\vspace{-0.5em}
		\caption{Overview of the soft wrapped haptic display design. (a) Detailed view of square-cell array implemented in the 1-DoF sleeve display. The thick lines indicate places where the LDPE plastic tube was heat-sealed. The sleeve is composed of 3 pouches taped together to form a sleeve with circumference $2{\pi}R$. The sleeve display is shown (b) deflated and (c) inflated. (d) Detailed view of the square-cell array implemented in the 1-DoF ring display. Grouping multiple individually-actuated pouches, placed side by side, forms a N-DoF wrapped haptic displays. A 3-DoF ring display is shown in two states: (e) deflated and (f) one of the DoF (center) is actuated, while the others are deflated.}
		\label{fig:device}
	\end{center}
	\vspace{-2.0em}
\end{figure*}

\subsection{Requirements}\label{Requirements}
While designing the wrapped haptic display concept we considered three key requirements to improve operation and the haptic sensation: low volume, fast inflation, and textured surface. First, the display should function without using large volumes of air or static materials, keeping the display flexible enough to easily wrap around objects like the robot arm.  Limiting the volume of air also aids in fast inflation and deflation, allowing faster changes in the produced signals. Additionally, we aimed to create an inflatable surface with textured tactile sensations. We believe that a textured surface helps users to quickly identify pressure changes in the display by exploring surface features. 

\subsection{Soft Haptic Display Concept}\label{concept}
To address these requirements, we use thin, heat-sealable, and inextensible materials that are formed into air-tight pouches and heat-sealed with patterns. 
The heat sealed patterns subdivide the bag, limiting the volume, adding texture, and keeping the surface flexible when inflated. The final design consists of an array of 2.54~cm square-shaped cells patterned into a low-density polyethylene (LDPE) plastic tube, sealed using a linear heat sealer (H-89 Foot-Operated Impulse Sealer, ULINE). A repeated and homogeneous pattern with gaps in the seals between cells allows for smooth and fast inflation. 
The square-array design is shown in Figure~\ref{fig:device} in two form factors. The dimensions and shape of the display can be varied to fit different applications and surfaces. A unit of a soft wrapped haptic display consists of one or more pouches attached to the same pressure source, forming a degree of freedom (1-DoF). Multiple degrees of freedom can be attached together to form an N degree of freedom (N-DoF) display. Given this general \textcolor{black}{description} of the soft haptic display, we will next describe the specific 1-DoF and 3-DoF displays used in testing.

\subsection{Large Surface Display}\label{subsec:1-dof}
The 1-DoF soft wrapped haptic display was made from a set of three connected pouches made from a 10.16~cm flat-width LDPE tube (S-5522, ULINE). The LDPE tube matched the length of one section of a UR-10 robotic arm (40.64 cm). 
The heat-sealed lines are 1.27 cm long, alternated in rows and columns to create the 2.54 cm-squares (Figure~\ref{fig:device}(a)). Through-wall straight connectors (5779K675, McMaster-Carr) were attached to one side of each bag strip to allow for individual inflation. 
The display was made of three bags taped together using viscoelastic adhesive tape (MD-9000, Marker Tape) to construct a sleeve that matched, and entirely wrapped, the cylindrical surface (Figure~\ref{fig:device}(a)-(c)).  The bags were connected using tee-adapters and inflated using a single pressure line (i.e. a 1-DoF soft wrapped haptic display). The 1-DoF soft wrapped haptic display can be inflated quickly; pressures above 1.5 psi (10.43 kPa) inflate in 0.86 seconds, the pressure can be changed from 1 to 3 psi (6.89 to 20.68 kPa) in 0.72 seconds, and deflate back to 1 psi in 0.18 seconds. The display can operate to a maximum of 3.5 psi (24.13 kPa). Above that pressure the heat-sealed edges begin to tear, producing leaks.

\subsection{Multi-Degree of Freedom Display}\label{subsec:n-dof}
We next increased the signal complexity while maintaining the design requirements by building on the 1-DoF design. We grouped multiple, individually-actuated pouches to form a N-DoF wrapped haptic display, as shown in Figure~\ref{fig:device}(d)-(f).  Each pouch consisted of a 2.54~cm flat-width LDPE tube (S-11155, ULINE), cut to fit the circumference ($2 \pi R$) of a segment on a robot arm and form a ring-shaped haptic display. Grommets were placed in the ends of the displays, and elastic bands tied the device around the cylindrical surface.
The pattern is modified from the 1-DoF displays to better fit the LDPE tubing. The 2.54~cm square cell grid was achieved by heat sealing 1.7~cm long lines across the length of the tube, alternating sides (Figure~\ref{fig:device}(d)). 
Silicon tubing (0.66cm OD) was attached to an end of the individual ring display to inflate. For the studies in Sections~\ref{sec:p2} and \ref{sec:vt2}, three ring displays were placed side by side. Separation between pouches (1.9~cm) is added to assist in making the identification of each DoF easier. 
Since the N-DoF display segments cover a smaller area, it is easier to mount them in different places of the robot arm. 
Additionally, since the width of these displays is smaller than the 1-Dof sleeve design, they have smaller volume when inflated and resist higher pressures, producing faster inflation/deflation speeds. These ring-shaped soft wrapped haptic displays can be inflated to pressures above 1.5 psi (10.43 kPa) in 0.55 seconds, and withstand a maximum of 5 psi (34.48 kPa). Switching from 1 to 3 psi (6.89 to 20.68 kPa) occured in 0.38 seconds, and the display deflate back to 1 psi in 0.12 seconds.


\subsection{Implementation} \label{subsec:implementation}

The haptic displays were mounted on cylindrical surfaces for the studies outlined in the following experiments, either sections of the robot arms or a PVC pipe acting as a stand-in. The mounting arrangements fixed the wrapped display in place, restricting it to 10\% contraction. The basic pneumatic control systems used to actuate the wrapped haptic displays consisted of: (1) a pressure regulator that supplied an electronically controlled pressurized-air supply and (2) a pressure release feature for deflating the displays. Two different pressure regulators were used. A pressure regulator with built in sensor and exhaust (QB3, Proportion-Air, McCordsville, Indiana) was used for the studies outlined in Sections \ref{sec:p1} and \ref{sec:p2} and was controlled using an Arduino Uno via MATLAB. 
A different pressure regulator (550-AID, ControlAir, Amherst, New Hampshire) was used for the remaining experiments, controlled using the UR-10's I/O controller (Section \ref{sec:vt1}) or an Arduino Uno (Section \ref{sec:vt2}). For this pressure regulator, the inflation pressure was measured using an electronic pressure sensor (015PGAA5, Honeywell Sensing, Gold Valley, Minnesota). 
If faster switching between inflation and full deflation is needed, on-off solenoid valves can be implemented. It is important to note that each 1-DoF device 
was connected to an individual pressure supply. For the case of the 3-DoF display, one can configure the device to effectively act as a 1-DoF device by connecting the individual rings to a single pressure, or have 3-DoF control if three pressure regulators are used.

%% file: study1-purdue.tex
\section{Measuring Human Perception of 1-DoF Wrapped Haptic Displays} \label{sec:p1}

Understanding the human sensory perception of the soft display, especially compared to rigid haptic displays, is essential in applying and controlling the wrapped haptic display. 
To that end, we first conducted a psychometric user study to measure the ability to distinguish display signals outside of the context of the target scenario. Participants physically interacted with the 1-DoF display and were asked to distinguish between pairs of pressures. We studied the user’s ability to differentiate inflation levels in the display to understand what pressure differences produce clear signals. \textcolor{black} {This experiment was previously featured in greater detail in our previous work~\cite{valdivia2022wrapped}}.

\subsection{Experiment Setup} \label{Exp Setup P}

\begin{figure}[t]
	\begin{center}
		\includegraphics[width=0.85\columnwidth]{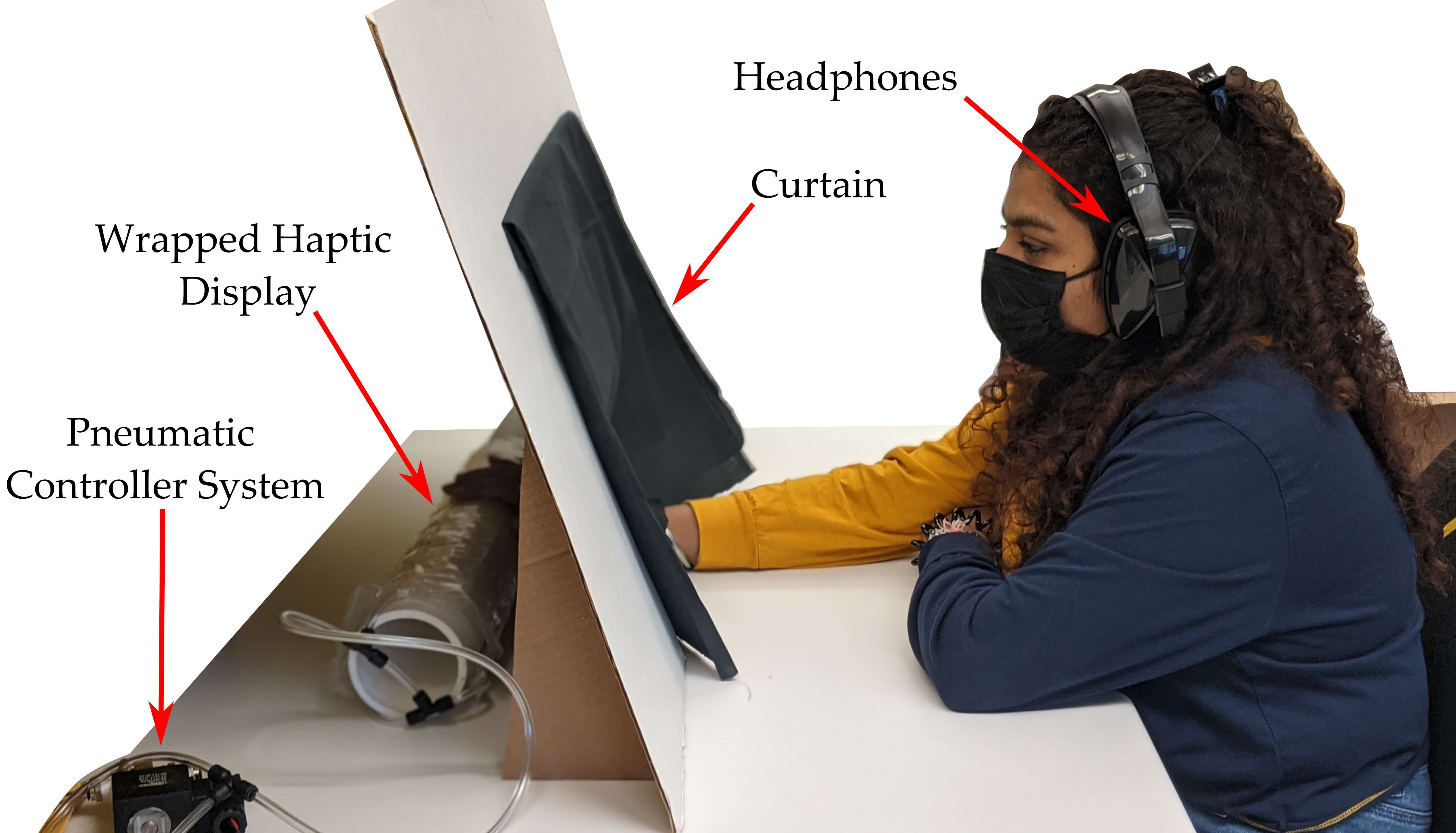}
		\vspace{-0.5em}
		\caption{Experimental Setup. The participants were instructed to sit at the desk right in front of the curtain and put on hearing protection headphones.}
		\label{fig:exp_setup}
	\end{center}
	\vspace{-2.0em}
\end{figure}

The 1-DoF inflatable haptic display was mounted on a PVC pipe matching the diameter of the UR-10 (Section~\ref{sec:vt1}), and the pipe was secured flat to the table. A curtain blocked the user’s vision, and users wore hearing protection to ensure the perception study focused on tactile sensations (Figure~\ref{fig:exp_setup}).

The study was conducted as a forced-choice comparison where participants were asked to identify the higher of two pressures. Pressures were shown in pairs (i.e., reference pressure, $P_o$, vs. test pressure, $P$), distinguished as ``Pressure 1'' and ``Pressure 2''. We selected 2 psi (13.79 kPa) as the reference pressure, and the test pressure values of 1.5, 1.75, 1.875, 2.0, 2.125, 2.25, and 2.5 psi (10.34, 12.07, 12.93, 13.79, 14.65, 15.51, and 17.93 kPa). These pressures are within a safe operating range for the display. We randomized the order in which the pairs were shown to the participant, as well as the order of reference and test pressure in each pair. As a note, in some pairs the reference pressure and test pressure were both 2.0 psi to measure bias in participants' choices  when guessing. 

The participants sat at the desk, and, before beginning the experiments, we demonstrated the display function, allowing participants to interact with it. Each experimental trial began by inflating the display to ``Pressure 1''. The participants were told to interact with the display for an unrestricted period of time and then release it. Then, the display was inflated to ``Pressure 2,'' and the participants were asked to interact again. Once they interacted with both pressures, we asked which one felt like a higher inflation pressure. The subjects were not told the correct answers. This procedure was repeated ten times each for the seven test pressures. After completing the interaction portion, the participants were given a post-experiment questionnaire asking about their overall study experience and their previous experiences with haptic technology, robotics, etc. The entire experiment took approximately 35 minutes, with an optional break. 

\begin{figure}[t]
	\begin{center}
		\includegraphics[width=0.89\columnwidth]{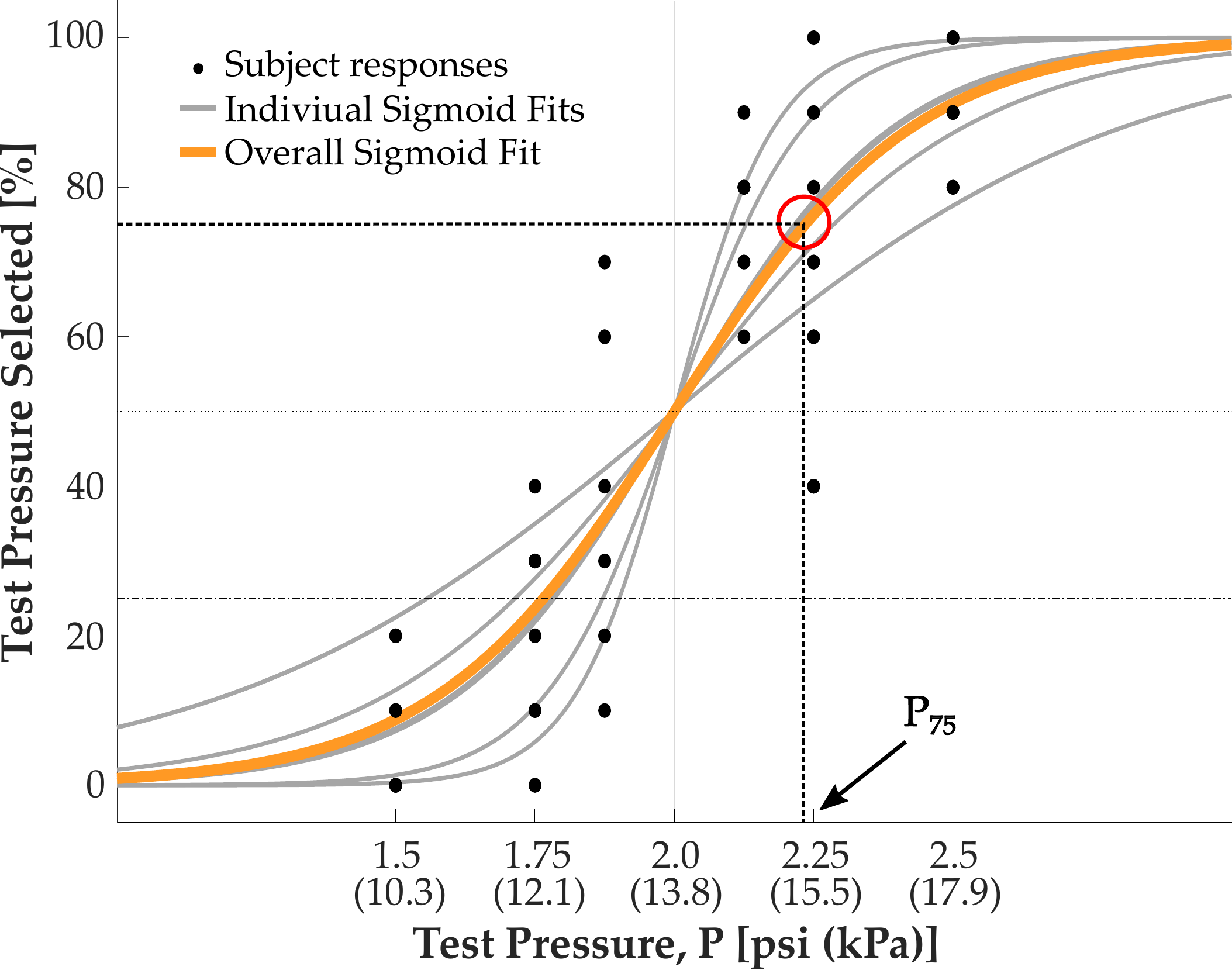}
		\vspace{-1em}
		\caption{\textcolor{black}{Raw data and sigmoid function fit for the collection of responses (orange) and each subject (grey). The percentages represent the proportion of times subjects selected the test pressure, $P$, as higher. The JNDs were calculated using the sigmoid function to solve for the pressure value corresponding to the 75\% threshold and subtracting it from the reference pressure. The dots represent percentages associated with individual subject responses. The k steepness factor for the overall sigmoid fit was 4.678, giving a JND of 0.235~psi. The individual steepness factors ranged 2.477-11.15, with JNDs varying between 0.099 and 0.444~psi (0.68-3.06~kPa).}}
		\label{fig:sigmoid}
		\vspace{-0.75em}
	\end{center}
\end{figure}


\subsection{Results}

\textcolor{black}{A total of $10$ participants ($4$ female, 1 non-binary, 5 male, average age $20.6$ years, age range $18-23$ years)} participated in this experiment after giving informed consent. Out of the group, $9$ participants were right-handed, and $1$ was left-handed. The Purdue Institutional Review Board approved the study protocols. Figure \ref{fig:sigmoid} shows the subjects' responses to the experiment. Each dot shows the percentage of times the test pressure was selected as higher when compared to the reference pressure. The just noticeable difference (JND) was calculated by first fitting a sigmoid function to the data:
\begin{equation}\label{sigmoid}
    q = \frac{100}{1+e^{-k(P-P_o)}}
\end{equation}
where $q$ is the modeled percentage of times the user chose the test pressure ($P$) as higher, $k$ is the steepness factor for fitting a sigmoid curve, $P$ is the test pressure, and $P_0$ is the reference pressure. 
Using this fit, the JNDs are calculated by finding the pressure value corresponding to the 75\% threshold, $P_{75}$, and subtracting the reference pressure, $P_0$:
\begin{equation}\label{JND}
    JND = P_{75}-P_o = -\frac{1}{k} ln\left(\frac{100}{75}-1 \right) 
\end{equation}
The sigmoid function was fit for each of the subjects, as well as for the collection of responses from all subjects.

\subsection{Analysis} \label{subsec:study1_analysis}

The experimental results show that the $k$ steepness factor for the overall sigmoid fit (the orange line in Figure~\ref{fig:sigmoid}) was 4.678, with 95\% confidence bounds between 3.605 and 5.751, giving a JND of 0.235~psi (1.62~kPa).
Individual JNDs ranged 0.099-0.444 psi (0.68-3.06 kPa).  The mean JND was defined as the mean of the values obtained for all participants, which was found to be 0.228 psi (1.57 kPa) with a standard deviation of 0.109 psi (0.75 kPa). The Weber fraction (WF), calculated as the ratio of the JND and the reference pressure, ranged between 4.9\% and 22.2\%, with a mean value of 11.4\%. Although there was no restriction on how the user could interact with the display, multiple users reported using active interaction to explore the display. This means that participants used reactive force sensing to explore the dynamics of inflation and determine how much pressure was used. Additionally, users reported mainly using their fingertips. Previous studies on fingertip psychophysics tests show similar results. Frediani and Carpi \cite{frediani2020tactile} conducted psychophysical tests for a fingertip-mounted pneumatic haptic display, reporting JNDs of 0.12-0.33 psi (0.8-2.3 kPa) for pressures between 0.58 and 2.90 psi (4 and 20 kPa), yielding a WF of 15\%. A study evaluating a haptic jamming display found fingertips WF to be 16\% ($\sigma$ = 7.4\%) and 14.3\% ($\sigma$ = 2.6\%) for stiffness and size perception, respectively \cite{genecov2014perception}. A different study testing stiffness perception for a rigid vibrotactile, fingertip-mounted haptic device reported WF between 17.7 and 29.9\% \cite{maereg2017wearable}. The JNDs and WFs obtained in this study show that our wrapped haptic display produced detectable signals and matched previously found psychometric baselines. 

	

As mentioned in Section \ref{Exp Setup P}, the reference pressure was shown against itself 10 times to measure subject's bias. Subjects overall showed unbiased behavior, choosing ``Pressure 1'' 45\% of the time and ``Pressure 2'' 55\% of the time. However, two subjects had a large preference for choosing ``Pressure 2'' (80\% of the time when guessing). These subjects also scored the highest WF, which may explain their higher bias when guessing compared to the complete participant pool. 

The qualitative data from the post-experiment questionnaire shows that, besides the participants already mentioned (who had the highest WF), no other participants struggled to identify the pressures. A majority of the participants ($7$ out of $10$) agreed that they could detect the differences and that they were sure about their answers. It is also worth noting that subjects with the highest performance reported dexterity-related skills, such as playing string musical instruments, piano, knitting, and American Sign Language proficiency.

This study shows that the sensations produced by our wrapped haptic display match the psychometric measures for other haptic devices. Both quantitative and qualitative results show that users were able to distinguish pressure changes over time without a specific task context. Overall, we demonstrated that the soft-wrapped haptic display can perform as well as other haptic devices in displaying tactile signals. 



\subsection{Follow-Up Study}
As timing became a significant factor during the later studies in Sections~\ref{sec:vt1}-\ref{sec:vt2}, a follow-up study was conducted, replicating the experimental procedure with the addition of a graphical user interface (GUI). The purpose of the GUI was to enable participants to control the pace of the experiment without the influence of the experimenter and to allow accurate recording of the time spent exploring each pressure. By evaluating time, we can better understand later result on timing and difficulty of interpreting haptic signals.

\textcolor{black}{A total of $12$ participants ($6$ female, $0$ non-binary, $6$ male, average age $21.9$ years, age range $21 - 23$ years)} participated in the follow-up experiment after providing informed consent. Due to technical difficulties in data collection, 2 participants were removed from the study. 1 additional participant was excluded from analysis as an outlier (performance equivalent to guessing). Of the remaining 9 participants, 7 were right-handed, and 2 were left-handed. 

The results show a JND of 0.279 psi (1.923 kPa). Individual JNDs ranged 0.114-0.674 psi (0.788-4.650 kPa), with a mean JND of 0.310 psi (2.136 kPa) and standard deviation of 0.173 psi (1.195 kPa). The WF ranged between 5.7\% and 33.7\%, with a mean value of 15.5\%, consistent with the initial study. 

Participants spent an average of 13.84~s on the first pressure ($\sigma$ = 7.323~s), and an average of 11.27~s on the second pressure ($\sigma$ = 5.746~s), for an average of 25.11 s per pair ($\sigma$ = 10.855~s). By one-way ANOVA, total time spent per pair was found to significantly impact correctness ($p$ = 0.024). Subjects spent significantly more time assessing the haptic device when answering incorrectly (26.84~s) than when answering correctly (24.56~s). Notably, mean time itself did not have a significant influence on overall accuracy ($p$ = 0.973).  
\vspace{-0.5em}

%% file: study2-vt.tex
\section{Applying Wrapped Haptic Displays to Communicate 1-DoF Robot Uncertainty} \label{sec:vt1}

So far we have studied the precision with which humans can perceive the 1-DoF wrapped haptic display. Next, we apply this display to convey robot learning from physical interactions. Section~\ref{sec:vt1} presents a condensed version of the robot experiments in \cite{valdivia2022wrapped}, excluding some details on tasks, metrics, and procedure. In this experiment participants kinesthetically taught a UR-10 robot arm to perform cleaning tasks. We applied an existing learning algorithm to measure the robot's uncertainty \cite{menda2019ensembledagger} and then conveyed that uncertainty back to the human in real-time. We highlight two key differences from the experiment in Section~\ref{sec:p1}: here the robot arm is \textit{moving} during interaction (i.e., the wrapped haptic display is not stationary), and the haptic display \textit{now conveys a specific signal} that the human must interpret and react to during interaction.

\begin{figure}[t]
	\begin{center}
		\includegraphics[width=0.95\columnwidth]{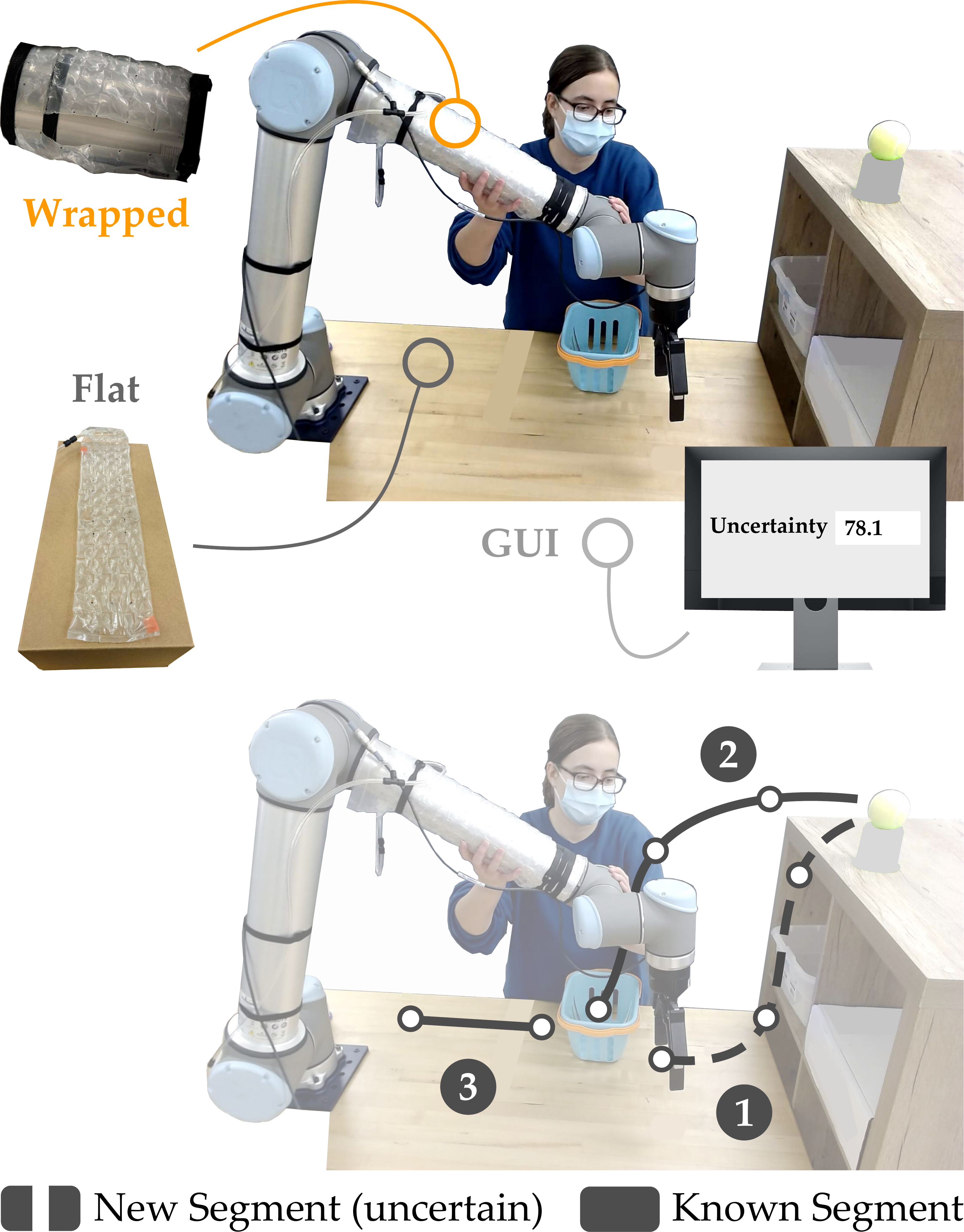}
		\caption{Participant kinesthetically teaching the robot arm the \textit{Cleaning} task. (Top) We compared our proposed approach (\textbf{Wrapped}) to two alternatives. \textbf{GUI} displayed the robot's uncertainty on a screen, while in \textbf{Flat} we placed the haptic display on table. (Bottom) We initialized the robot with data from known segments. During their first demonstration the human attempted to identify the region where the robot was uncertain (i.e., the new segment). The human then gave a second demonstration where they only guided the robot through the region(s) where they thought it was uncertain.}
		\label{fig:setup_study2}
	\end{center}
	\vspace{-1.5em}
\end{figure}

\p{Independent Variables} We compared three different types of feedback (see \fig{setup_study2}): 
\begin{itemize}
    \item A graphical user interface (\textbf{GUI}) that displayed the robot's uncertainty on a computer monitor.
    \item Our soft haptic display placed \textbf{Flat} on the table.
    \item Our proposed approach where we \textbf{Wrapped} the haptic display around the robot arm.
\end{itemize}
All three types of feedback showed the same information but used different modalities. In the \textbf{GUI} baseline we displayed uncertainty on a computer screen in front of the user. Here uncertainty was shown as a percentage, where numbers close to $100\%$ indicated that the robot was uncertain at the current state. The \textbf{Flat} and \textbf{Wrapped} interfaces used the 1-DoF soft haptic display from Section \ref{Haptic Display and Design}. Uncertainty was linearly scaled from $1-3$ psi ($6.89 - 20.68$ kPa). Here $1$ psi (deflated bags) corresponded to $0\%$ uncertainty and $3$ psi (inflated bags) corresponded to $100\%$ uncertainty. The \textbf{Flat} haptic display was placed in a designated area next to the human, such that participants could periodically touch it during the experiment.

\begin{figure*}[t]
	\begin{center}
		\includegraphics[width=2.0\columnwidth]{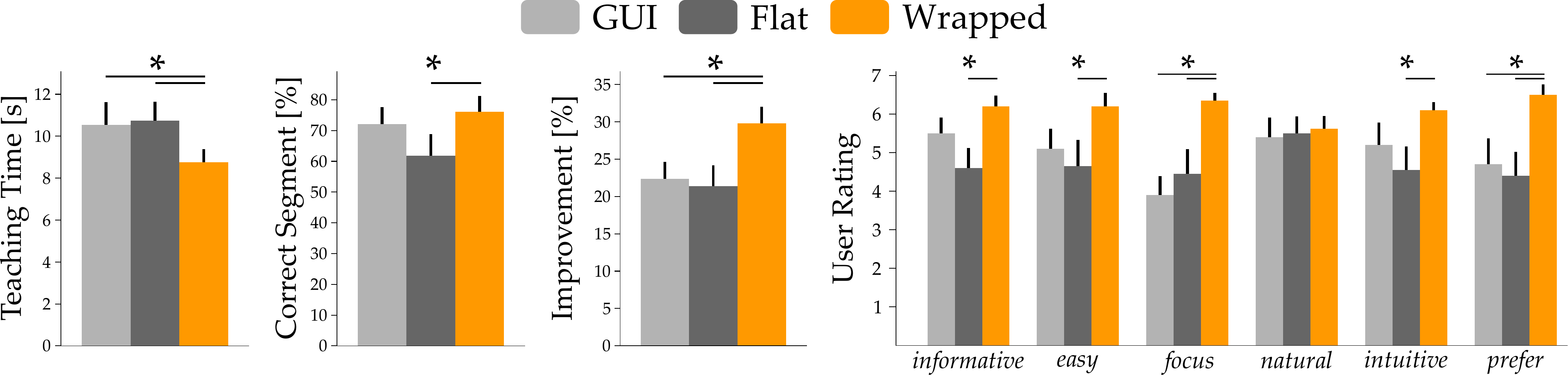}
		\caption{Objective and subjective results when communicating 1-DoF robot uncertainty in real-time with \textbf{GUI}, \textbf{Flat}, and \textbf{Wrapped} feedback. Participants taught the robot three tasks; we here report the aggregated results across tasks. Error bars show standard error of the mean (SEM), and $*$ indicates statistically significant comparisons ($p < .05$). (Left) Wrapping the haptic display around the robot arm caused participants to spend less time teaching the robot, focused their teaching on regions where the robot was uncertain and improved the robot's understanding of the task after the human's demonstration. (Right) Participants thought that the wrapped display best enabled them to focus on the task, and they preferred this feedback type to the alternatives.}
		\label{fig:results_study2}
	\end{center}
	\vspace{-1.5em}
\end{figure*}

\p{Experimental Setup} Participants completed three different tasks with each of the three feedback conditions (i.e., nine total trials). Tasks involved pushing, grasping, and moving objects around a table and drawers. \fig{setup_study2} shows an example task.

Before conducting any experiments we first initialized the robot's uncertainty. We collected five expert demonstrations of each task and trained the robot with a behavior cloning approach \cite{menda2019ensembledagger}. This approach outputs the robot's uncertainty at each state (i.e., uncertainty was a function of the robot's joint position). We purposely \textit{removed} segments of the expert's demonstrations from the training set: specifically, we trained the robot without showing it how to perform either the first segment or the last segment of the task. As a result, when participants interacted with the robot, the robot was uncertain about either the start or the end of the task.

For each trial the participant provided \textit{two demonstrations}. First, the participant kinesthetically guided the robot throughout the entire task while receiving real-time feedback from \textbf{GUI}, \textbf{Flat}, or \textbf{Wrapped}. Based on this feedback, the participant attempted to identify the region of the task where the robot was uncertain (and needed additional teaching). During the second demonstration, the human \textit{only taught the segment} of the task where they believed the robot was \textit{uncertain} (i.e., the region they identified in the first demonstration). If the feedback is effective, participants should only reteach segments where the robot is confused without repeating parts of the task that the robot already knows.

\p{Participants} \textcolor{black}{We recruited $10$ participants from the Virginia Tech community to take part in our study ($5$ female, $0$ non-binary, $5$ male, average age $22.9$, age range $19 - 26$ years)}.


\p{Dependent Measures}
To measure how the robot's feedback affected the human's teaching, we focused on the \textit{second demonstration} (i.e., the demonstration where users retaught the uncertain part of the task). We recorded the time users spent on this second demonstration (\textit{Teaching Time}) and the percentage of this second demonstration that overlapped with the segment where the robot was actually uncertain (\textit{Correct Segment}). Offline, we retrained the robot using the participant's second demonstration. We then measured the percentage reduction in uncertainty due to the user's demonstration (\textit{Improvement}). Finally, we also measured how users \textit{subjectively perceived} each feedback method using a 7-point Likert scale survey.

\p{Hypotheses}
We had two hypotheses for this user study:
\begin{displayquote}
    \textbf{H1.} \emph{Participants will most efficiently teach the robot with wrapped haptic displays.} 
\end{displayquote}
\begin{displayquote}
    \textbf{H2.} \emph{Participants will subjectively prefer our  wrapped haptic display over other methods.} 
\end{displayquote}

\p{Results} We summarize our aggregated results in \fig{results_study2}.

We first ran a repeated measures ANOVA, and found that the robot's feedback type had a statistically significant effect on \textit{Teaching Time}, \textit{Correct Segment}, and \textit{Improvement}. Post hoc analysis revealed that participants spent less time teaching the robot with \textbf{Wrapped} than with either \textbf{GUI} or \textbf{Flat} ($p < .05$). Participants also better focused their teaching on the region where the robot was actually uncertain: \textbf{Wrapped} resulted in a higher \textit{Correct Segment} than \textbf{Flat} ($p < .05$). However, here the differences between \textbf{Wrapped} and \textbf{GUI} were not statistically significant ($p=.287$).

Recall that \textit{Improvement} captures how much more confident the robot is about the task after the participant's demonstration. This metric is especially important: we want to enable humans to teach robots efficiently, and \textit{Improvement} quantifies how much the robot learned from the human's teaching. We found that the robot's confidence improved the most in the \textbf{Wrapped} condition as compared to either \textbf{GUI} or \textbf{Flat} ($p < .05$). Overall, these results support \textbf{H1}: when users get real-time feedback from a haptic display wrapped around the robot arm, they provide shorter duration kinesthetic demonstrations that more precisely hone in on the robot's uncertainty and efficiently correct the robot.

We next analyzed our Likert scale survey to understand how users perceived each type of feedback. After confirming that our six scales were reliable (using Cronbach's $\alpha$), we grouped these scales into combined scores and ran a one-way repeated measures ANOVA on each resulting score. Post hoc analysis showed that participants thought that \textbf{Wrapped} was more informative, easier to interact with, less distracting, and more intuitive than either one or both of the alternatives ($p < .05$). Participants also indicated that they preferred \textbf{Wrapped} over \textbf{GUI} and \textbf{Flat}. When explaining this preference, one participant said, \textit{``I definitely prefer \textbf{Wrapped} over other methods. I was able to clearly focus and the other methods were distracting.''}. Our subjective results support \textbf{H2}, and indicate that users perceived wrapped haptic displays as preferable when compared to alternatives like visual interfaces. \textcolor{black}{We also note that the overall results from videos, user ratings, and teaching time indicate that participants were able to detect \textbf{Wrapped} feedback during their kinesthetic demonstrations, i.e., in \textbf{Wrapped} users did not need to stop moving, explore the pouches, and then resume their demonstration.}


%% file: study3-purdue.tex
\section{Measuring Human Perception of 3-DoF Wrapped Haptic Displays} \label{sec:p2}

Having explored the human perception and application of the 1-DoF wrapped haptic display in the shape of a sleeve, we next pursue a study that will help us understand how the spatial distribution of displays affects the perception of both 1-DoF and multiple-DoF soft haptic displays. Both temporally and spatially varying signals can help us add complexity when we need to communicate multiple haptic signals within the space that a human might contact. We also seek to understand how the spatial distribution of signals might affect the effectiveness of the display in terms of accuracy of identification and the time needed to identify signals. To do so, we conducted a user study to measure the ability to distinguish haptic signals in different spatial distributions and outside of the context of the target application scenario. We select pressure levels considering the psychometric baselines (JNDs) obtained in Section~\ref{sec:p1}, and designed a study in which participants physically interacted with 3-DoF displays. The displays were arranged in two ways: (1) a 3-DoF ring display placed in a single location, and (2) three 1-DoF displays made up of three rings each (by interconnecting the individual rings) and placed at three different locations. We called these arrangements \textbf{Global} (for the 3-DoF display), meaning all information was available at the single point of contact, so it would be "globally" available, and \textbf{Local} (for the three 1-DoF displays), meaning the information for each degree of freedom was only available locally. In each display, the user was asked to identify the signal with the highest pressure out of the three, and we hypothesized that the distribution of the signals (whether three in a single location or spread over a distance) would affect performance. As a note, these same methods are later used in the experiment in Section~\ref{sec:vt2}, but there three of the \textbf{Global} displays are used instead of one to keep the total area of the display on the robot constant. This also allows different users to contact the robot arm at different locations based on preference while still receiving the same feedback.

\begin{figure}[t]
	\begin{center}
		\includegraphics[width=0.8\columnwidth]{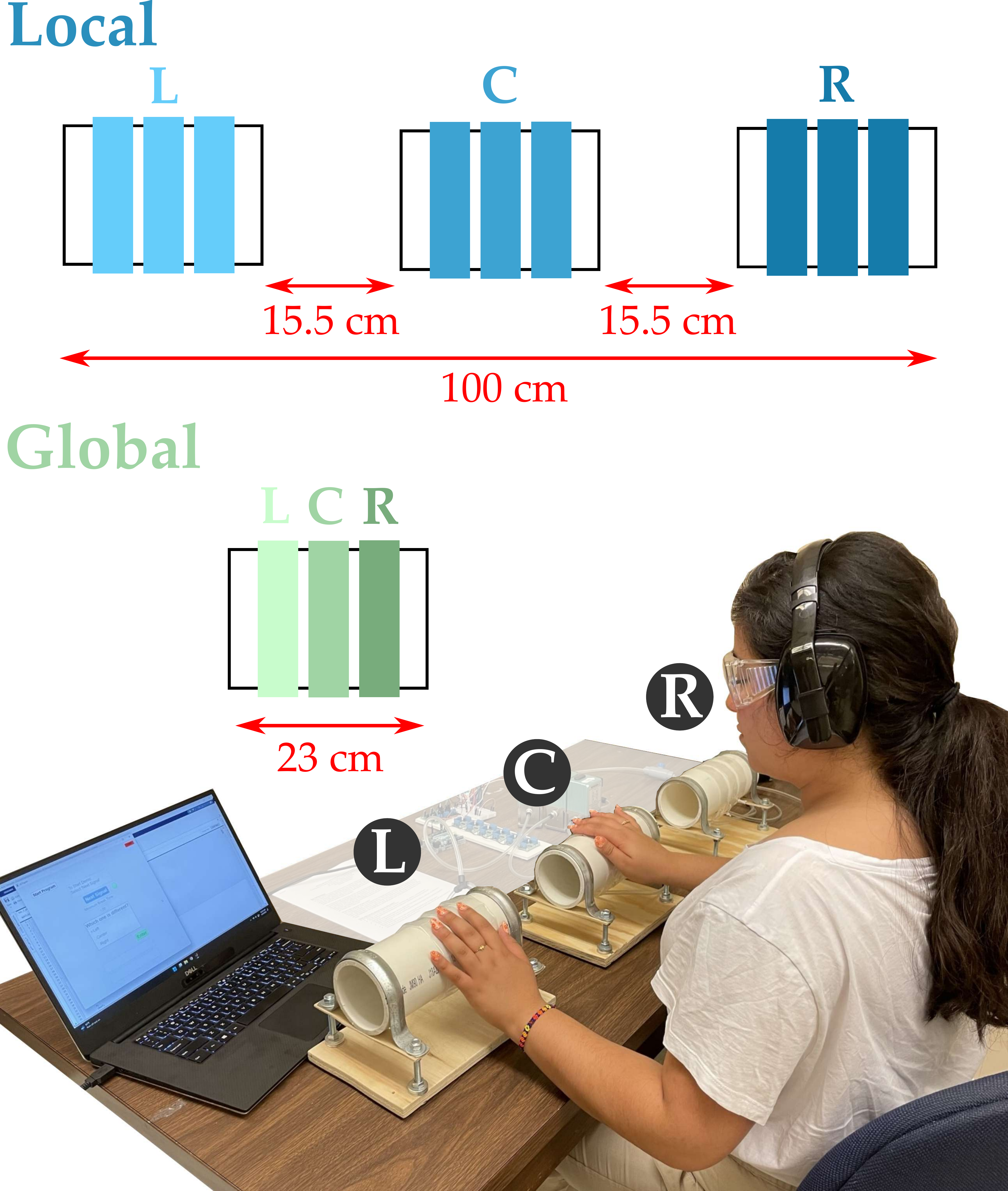}
		\vspace{-0.5em}
		\caption{Experimental Setup. (Top) The \textbf{Local} setup consists of three sets of 3-DoF displays configured as 1-DoF each, with a separation in between each. (Middle) The \textbf{Global} setup consists of a single 3-DoF display. Both methods essentially have 3-DoF, but the difference is the spatial distribution of each of the DoF. The three DoF were named \textit{Left}, \textit{Center}, and \textit{Right}, for both methods. (Bottom) Participants were instructed to sit in front of the setup; here, we show a participant interacting with the \textbf{Local} setup.}
		\label{fig:exp_setup3}
	\end{center}
	\vspace{-2.0em}
\end{figure}

\subsection{Experiment Setup and Procedure} \label{Exp Setup P2}
The 3-DoF wrapped haptic displays were mounted on passive stand-ins. For the \textbf{Local} method, three stand-ins with three ring displays configured as 1-DoF displays each were placed on the table, with a separation in between each. For the \textbf{Global} method, a single stand-in with a 3-DoF display was used. Both methods essentially have 3-DoF, but the difference is the spatial distribution of each of the degrees of freedom; for \textbf{Global}, all signals are located in a small space, while for \textbf{Local} the signals are distributed in a 1~m space. The three degrees of freedom were named \textit{Left}, \textit{Center}, and \textit{Right}, for both methods. The setups are illustrated in Figure~\ref{fig:exp_setup3}. Participants were instructed to wear hearing protection and safety glasses during the study. The task was to identify which of the signals, \textit{Left}, \textit{Center}, or \textit{Right}, was inflated to the higher pressure. Two of the degrees of freedom were inflated to a reference pressure $P_o$ (2psi) and one to a high pressure $P_H$ (2.75 psi). Subjects were not told that two degrees of freedom had the same pressure, they were just instructed to identify the one inflated to the different pressure. We selected the $P_o$ and $P_H$ values based on the findings of the previous pyschophysics study and taking into consideration that there is an increase in the complexity of haptic signals for this new study. As reported in Section~\ref{subsec:study1_analysis}, the average JND found in the previous study was 0.228~psi. However, some of the participants had JNDs almost double of the mean. With that in mind, we determined that a pressure difference between the signals of $\Delta$P = 0.75~psi was large enough so that we could be sure all subjects would perform to an adequate level in this study. 

Each DoF (\textit{Left}, \textit{Center}, and \textit{Right}) was rendered to the participant as the $P_H$ a total of 16 times each, for a total of 48 trials. The process was performed for both \textbf{Global} and \textbf{Local} methods. Half of the participants completed the procedure with \textbf{Global} first and the other half completed \textbf{Local} first. The study was as follows. Participants were instructed to sit at the desk in front of the arrangements. They interacted with a GUI developed in MATLAB to navigate through the study. The GUI first guided the participants through a demo to demonstrate the study procedure. The GUI showed a red light that would turn green to indicate when the participant was allowed to touch the displays. For each of trial, the GUI asked the participant to click a ``Next'' button to continue. Once clicked, the light would turn green once the displays reached their corresponding steady-state pressures. The participants were allowed to touch the displays for an unrestricted period of time and they could explore the displays using any method, including using both hands if desired. During this time, the GUI displayed the question \textit{``Which one has the different pressure?''}, and showed options for \textit{Left}, \textit{Center}, and \textit{Right}. After the participants selected an option and confirmed by clicking an ``Enter'' button, the GUI showed whether they were correct and, if incorrect, what the right answer was. Note that the GUI was configured to measure the participants' response time; an internal timer would start when the light turned green and would stop when the participants answered the question. To continue with the next trial, the participants then \textcolor{black}{had to press ``Next.''} The procedure was repeated until 48 trials were completed for the first method and then for the second method. Participants were offered a break halfway through each method and another break in between methods. After completing the interaction portion of the experiment, participants answered a post-experiment questionnaire. The questionnaire asked about how distinguishable the signals were, if they were often unsure about their answers and if they were increasingly confident about their answers as the study progressed. We also asked about the overall experience during the study (clarity of instructions, sense of safety during the experiment) and about their previous experiences and familiarity with haptic technology, robotics, etc. The study was 45 minutes long. 

\begin{figure}[t]
	\begin{center}
		\includegraphics[width=1.0\columnwidth]{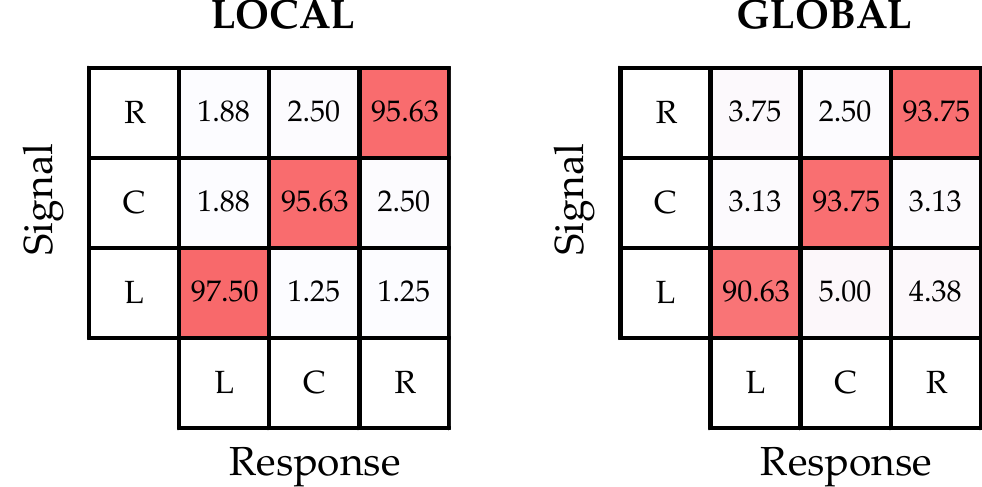}
		\caption{Confusion matrices showing the mean accuracy for each signal rendered (\textit{Left}, \textit{Center}, \textit{Right}) in both methods (\textbf{Local} and \textbf{Global}).}
		\label{fig:matrices}
	\end{center}
	\vspace{-2em}
\end{figure}

\begin{figure*}[t]
	\begin{center}
		\includegraphics[width=1.9\columnwidth]{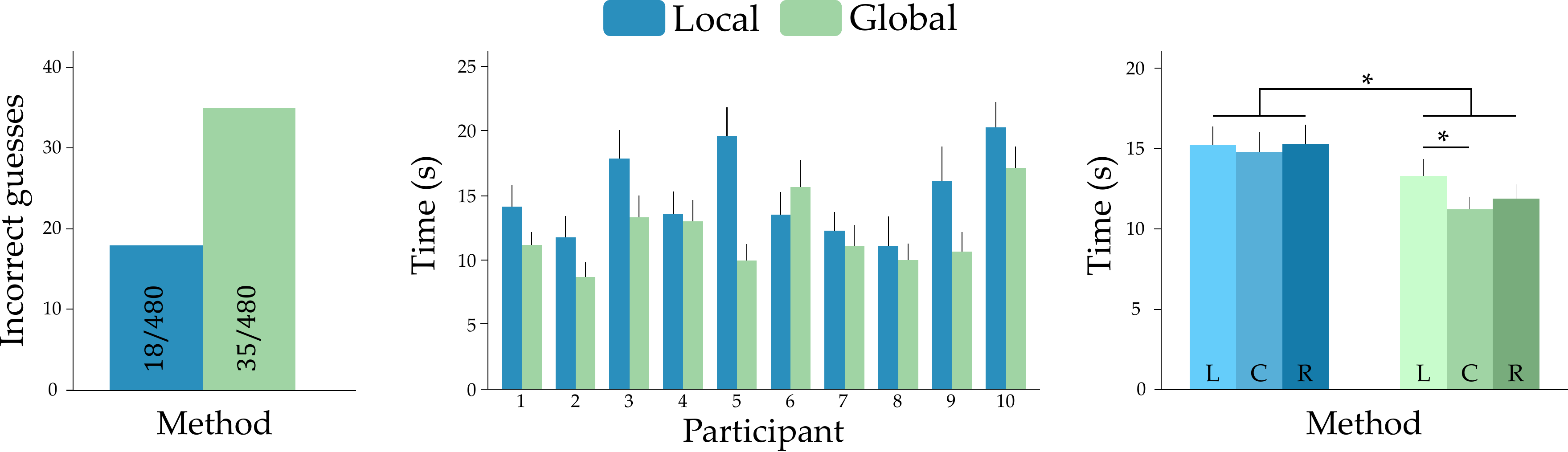}
		\vspace{-0.5em}
		\caption{Experimental results. (Left) Count of incorrect guesses for each of the methods. A Wilcoxon signed-rank test showed that there is a significant association between participants' accuracy and methods ($Z$~=~-2.335, $p$~<~.05). (Center) Mean response time of individual participants for each method. Nine out of ten participants had a higher response time using the \textbf{Local} method. (Right) Mean response time for both methods, displayed by signal type (\textit{Left}, \textit{Center}, \textit{Right}). Signal type had a statistically significant effect on response time ($p$~=~.047) but to a lesser extent than the method type ($p$~<~.001). For \textbf{Global}, participants spent more time responding the question when the \textit{Left} signal was the highest pressure than when it was \textit{Center} ($p$~<~.01) or \textit{Right} ($p$~=~.078).}
		\label{fig:results_study3}
	\end{center}
	\vspace{-2.0em}
\end{figure*}

\subsection{Results}
\textcolor{black}{We recruited $10$ participants ($4$ female, $0$ non-binary, $6$ male, average age $22.1$ years, age range $19 - 25$ years)} from the Purdue community. All participants completed the study after giving informed consent. The Purdue Institutional Review Board approved the study protocols (IRB \#2021-1283). In the group, 9 participants were right-handed; one was left-handed. 

The confusion matrices in Figure~\ref{fig:matrices} summarize the accuracy of participants. Overall, participants’ accuracy was higher for the \textbf{Local} method (average $\Bar{x}$~=~96.25\%, standard deviation $\sigma$~=~3.88) than \textbf{Global} ($\Bar{x}$~=~92.71\%, $\sigma$~=~7.17). Participants spent an average of 15.09s ($\sigma$~=~7.55) using the \textbf{Local} method, and 12.12s ($\sigma$~=~5.877) for \textbf{Global}. Interestingly, looking at the complete pool of participants' responses (whether global or local), we found that participants had a greater response time when they responded incorrectly ($\Bar{x}$~=~16.89s, $\sigma$~=~7.68s) than when they answered correctly ($\Bar{x}$~=~13.41s, $\sigma$~=~6.83). Figure~\ref{fig:results_study3} shows the average time spent by each participant for both \textbf{Local} and \textbf{Global} methods. 

\subsection{Analysis}
The two quantitative measures that we used to understand the results are \textit{Accuracy} and \textit{Response Time}. To further analyze the accuracy of participants, a Wilcoxon signed-rank test was conducted to understand the relation between accuracy and the methods used. The results showed that there is a significant association between participants' accuracy and methods ($Z$~=~-2.335, $p$~<~.05). This means that although participants responded faster to the task while using \textbf{Global} as shown by the mean response time values, participants were not as accurate at detecting the higher pressure as when they were using \textbf{Local}. Figure~\ref{fig:results_study3} shows the count of incorrect guesses for both \textbf{Local} and \textbf{Global} methods. Another Wilcoxon test was conducted to determine whether the order in which the experiments were conducted (\textbf{Local} first, then \textbf{Global}, or vice-versa) affected subjects' accuracy. The results showed that there was no significant association ($Z$~=~-0.143, $p$~=~.886), suggesting that subjects did not benefit from learning to improve their accuracy for the second half of the study. 

To analyze response time, we used a one-way repeated measures ANOVA. We found that the method type had a statistically significant effect on response time. Post hoc analysis revealed that participants spent less time identifying the target signal with \textbf{Global} as compared to \textbf{Local} ($p$~<~.001). This observation matches the mean values for the response time previously mentioned and also the mean response time for each participant, as shown in Figure~\ref{fig:results_study3}. Nine out of ten participants spent more time using \textbf{Local} compared to \textbf{Global}. We also found that the rendered signal (whether \textit{Left}, \textit{Center}, or \textit{Right}) had a statistically significant effect on answering time ($p$~=~.047) but with a smaller effect size than the method type. Data shows that while using \textbf{Global}, participants spent more time responding when the \textit{Left} signal was the highest pressure than when it was \textit{Center} ($p$~<~.01) or \textit{Right} ($p$~=~.078). For \textbf{Local}, we did not find any statistically significant distinction between signals and their mean response time. These results can be observed in Figure~\ref{fig:results_study3}, where we show the response time of participants for each signal type (\textit{Left}, \textit{Center}, \textit{Right}) when using \textbf{Local} and \textbf{Global} methods.

This study shows that the spatial distribution of haptic displays is an important factor to consider since it has an effect on both accuracy of detection and response time. Using the psychometric measures found in the previous study, we showed that participants were better able to identify the highest pressure out of a set of three when the signals were spatially distributed (\textbf{Local}) than when the signals were condensed in a smaller space (\textbf{Global}). However, the response time for the spatially distributed signals was higher; this makes sense because participants moved around a larger space to interact with the places where the haptic signals were located. Participants reported the pressure differences were detectable, they were sure about their answers throughout the experiment, and that they felt safe interacting with the displays. Some participants mentioned that during the \textbf{Local} portion of the experiment, they wished they could place the displays together to make the exercise easier; this suggests that users consciously thought that having displays dispersed in different locations was an inconvenience, even though the results show participants were slightly more accurate with this method than with \textbf{Global}. To summarize, the \textbf{Global} method had the faster response time, but \textbf{Local} had the higher accuracy. These observations show the trade-off between response time and accuracy when we increase the complexity of haptic signals in a smaller space or distribute them in a larger space.

%% file: study4-vt.tex
\section{Using Multi-DoF Wrapped Haptic Displays to Communicate 3-DoF Robot Learning} \label{sec:vt2}

In Section~\ref{sec:vt1} we demonstrated that robot arms can leverage a haptic display to communicate with human teachers. However, this haptic device only had $1$-DoF: the same pressure was rendered along the entire robot arm. One degree-of-freedom is sufficient when the robot learner wants to convey whether or not it is uncertain --- but what if the robot needs to communicate more complicated feedback? For instance, the robot may want to indicate \textit{what} it is confused about or \textit{how} the human teacher could improve their demonstrations.

In our final user study we wrap multiple $3$-DoF haptic displays around a Franka Emika robot arm. Participants physically teach the robot to perform a mock welding task, and the robot applies multi-dimensional feedback to indicate what aspects of the task the human teacher must emphasize. 
Overall, our goal is to compare the two different feedback distributions shown in Section~\ref{sec:p2} and understand how they impact the human's physical demonstrations. Remember that we are wrapping haptic displays along the robot arm. One option is to \textit{localize} different signals to different parts of the arm, such that the place where the bags inflate helps indicate and remind users what the robot is uncertain about. Our second option is to \textit{distribute} all three signals along the entire arm; here the human perceives the same haptic rendering no matter where they grasp the robot. In this user study we explore how human teachers perceive and leverage multiple displays that use both feedback layouts.

\begin{figure*}[ht!]
	\begin{center}
		\includegraphics[width=2.0\columnwidth]{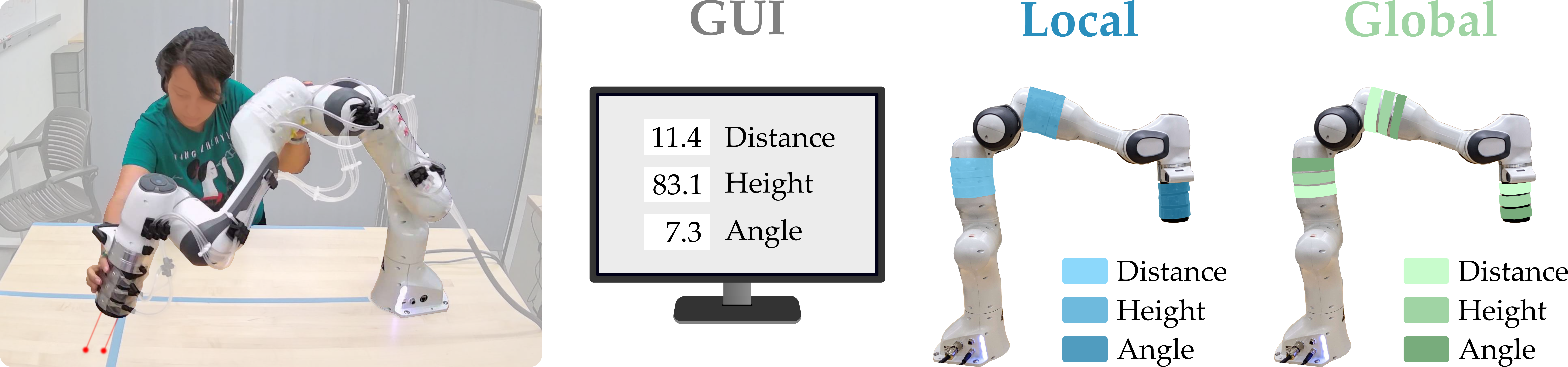}
		\caption{Experimental setup and independent variables for the user study from Section~\ref{sec:vt2}. (Left) Participants physically demonstrated a mock welding task to a Franka Emika robot arm. We mounted two lasers to the robot's end-effector: the robot prompted human teachers to keep the end-effector and lasers at the correct distance, height, and orientation. (Right) The robot indicated which feature(s) it needed help with using three different feedback modalities: \textbf{GUI}, \textbf{Local}, and \textbf{Global}. For \textbf{GUI} the robot printed its percentage uncertainty about each feature on a computer monitor placed in front of the workstation. Both \textbf{Local} and \textbf{Global} leveraged our wrapped haptic displays. In \textbf{Local} we attached three $1$-DoF displays, and the location of the display indicated the desired feature. By contrast, in \textbf{Global} we used three $3$-DoF displays such that each row of the displays corresponded to a separate feature.}
		\label{fig:setup_study4}
	\end{center}
	\vspace{-1.5em}
\end{figure*}

\p{Independent Variables} Participants kinethetically guided the robot arm through a mock welding task. The robot displayed feedback in real-time to guide the human through the task. We compared three different types of feedback for communicating when the robot was uncertain and what motions it needed the human teacher to emphasize (see \fig{setup_study4}):
\begin{itemize}
    \item A \textbf{GUI} baseline where the robot showed its numerical uncertainty on a computer monitor.
    \item Three $1$-DoF wrapped haptic displays with signals localized to different regions of the robot arm (\textbf{Local})
    \item Three $3$-DoF wrapped haptic displays with signals distributed across the entire robot arm (\textbf{Global})
\end{itemize}
All conditions provided the same information to the participants. Similar to Section~\ref{sec:vt1}, in \textbf{GUI} the robot displayed its uncertainties as a percentage: values close to $100\%$ meant that the robot needed assistance. For \textbf{Local} and \textbf{Global} we actuated three separate wrapped haptic displays with pressures between $1 - 3$ psi ($6.89 - 20.68$ kPa). In \textbf{Local} each location of the haptic display had a single pressure signal; i.e., bags at the end-effector were one pressure, bags at the base of the arm were another pressure, and bags in the middle of the arm were a third pressure. In \textbf{Global} each haptic display location rendered all three of the potentially different pressures using three independent degrees of freedom, and all \textbf{Global} displays rendered those same three pressures --- participants could feel the same feedback at the base, middle, and end of the robot arm. \textbf{GUI}, \textbf{Local}, and \textbf{Global} each provided a total of $3$-DoF feedback. We emphasize that with \textbf{Global} participants had to discern which segments of the $3$-DoF haptic display were inflated, while with \textbf{Local} participants needed to determine at which parts of the robot arm the haptic displays were inflated. 

\p{Experimental Setup} \textcolor{black}{Participants physically interacted with a $7$-DoF robot arm (Franka Emika) to complete a mock welding task (see \fig{setup_study4}). Recall that users interacted with a UR10 robot in Section~\ref{sec:vt1} --- from the user's perspective, the Franka Emika robot is smaller, has one more joint, and is easier to backdrive.} We mounted lasers to the robot's end-effector: participants kinesthetically guided the robot across a table while the lasers marked where the robot was ``welding.''

The welding task consisted of three features: how close the end-effector was to the edge of the table, the end-effector's height from the table, and the orientation of the end-effector. When the task started participants would guide the robot arm towards the fixed goal position. \textcolor{black}{As they moved, the robot would leverage its feedback to notify the human \textit{which feature} they needed to emphasize. For example, during the first third of the task the robot may prompt the human to keep the lasers close to the table; in the middle of the task the human should move the end-effector to the table edge; and during the final third of the task the human might need to align the robot's orientation.} Participants had to dynamically determine \textit{what} feature the robot currently needed help with and then \textit{modify} their motion to emphasize that feature. Note that the robot asked for assistance with all three features at different segments of the task --- we randomized these segments so that participants could not anticipate the robot's feedback.

\p{Participants and Procedure} \textcolor{black}{We recruited $12$ participants ($5$ female, $0$ non-binary, $7$ male, average age $28$, age range $19 - 35$ years) from the Virginia Tech community.} All participants provided informed written consent consistent with university guidelines (IRB \# $20$-$755$). None of the participants for this study took part in the previous study from Section~\ref{sec:vt1}. Three of the twelve participants reported that they had physically interacted with robot arms before.

Each user completed the welding task four times. First, we asked users to demonstrate the task without any feedback from the robot. We used this initial demonstration as a baseline to measure their improvement. Next, participants completed the welding task with \textbf{GUI}, \textbf{Local}, and \textbf{Global}. We counterbalanced the order of these feedback conditions: four participants started with \textbf{GUI}, four participants started with \textbf{Local}, and four participants started with \textbf{Global}.

\begin{figure*}[t]
	\begin{center}
		\includegraphics[width=1.9\columnwidth]{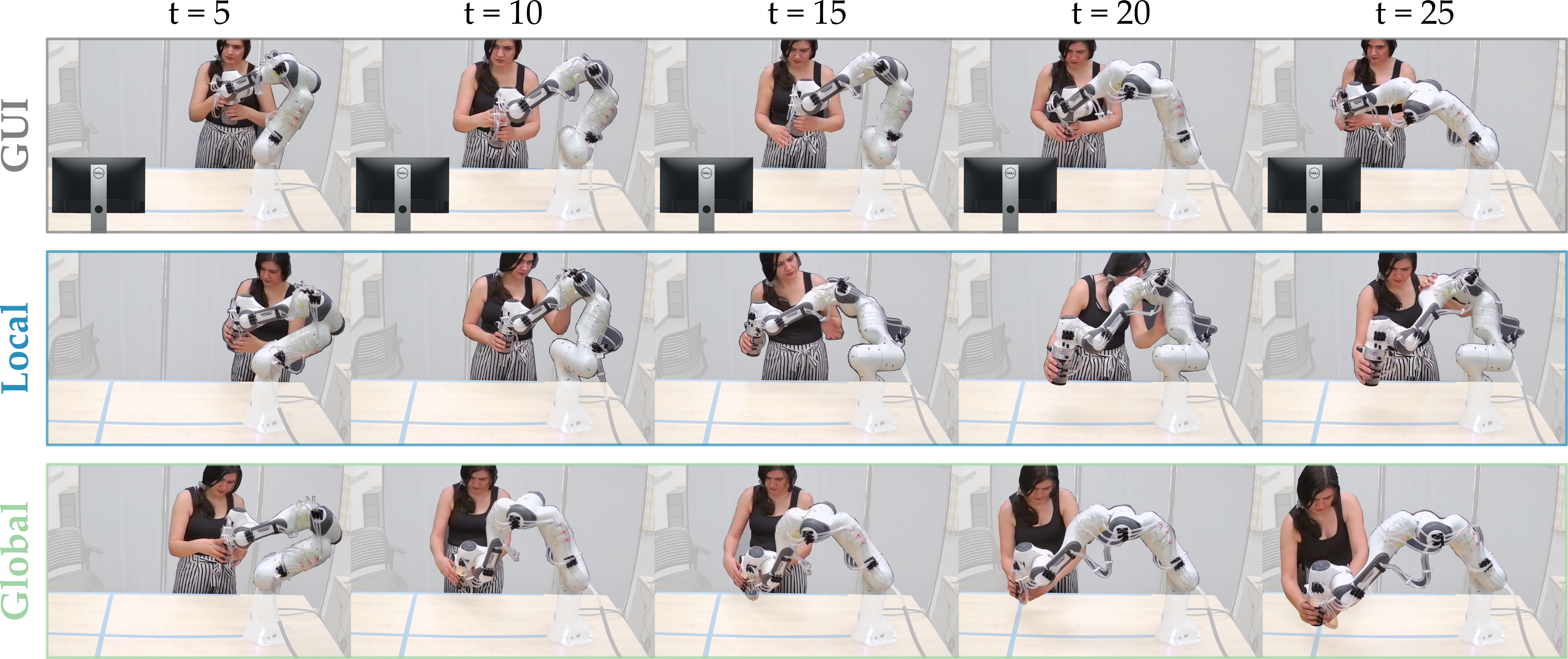}
		\caption{Participant teaching the welding task with \textbf{GUI}, \textbf{Local}, or \textbf{Global}. We show task progress in $5$ second intervals. (Top) With \textbf{GUI} users needed to look at the computer monitor to obtain feedback. The monitor is placed on the near side of the table: this participant is looking at the \textbf{GUI} at times $t=5$, $t=10$, and $t=25$ seconds. (Middle) With \textbf{Local}, participants must move their hands --- and change their grasp --- to sense the different wrapped displays. This participant keeps one hand on the end-effector, and then moves their other hand between the haptic displays at the middle and base of the robot arm. (Bottom) Finally, with \textbf{Global} the participants receive feedback through $3$-DoF Haptic displays. \textbf{Global} helped this user remain focused on the task: notice that they are continually looking at the welding task, and keep both hands on the end-effector (where a $3$-DoF haptic display is located).}
		\label{fig:example_study4}
	\end{center}
	\vspace{-1.5em}
\end{figure*}

\p{Dependent Measures -- Objective} \textcolor{black}{We measured the total time it took for participants to demonstrate the welding task (\textit{Teaching Time}). We also measured the \textit{Improvement} between the human's initial demonstration and their demonstration under each feedback condition. Let $f(\xi) \in \mathbb{R}^k$ be the feature counts along trajectory $\xi$, e.g., the distance, height, and angle. We define $e(\xi) = \| f(\xi) - f(\xi^*)\|^2$ as the error between the human's demonstration $\xi$ and the ideal trajectory $\xi^*$. \textit{Improvement} captures how this error changes after receiving robot feedback: $\big(e(\xi_{initial}) - e(\xi)\big)/e_{max} \cdot 100$, where $e_{max} = \| f(\emptyset) - f(\xi^*)\|^2$ is a normalizer (i.e., the error when the human does not provide any demonstration). \textit{Improvement} captures the percentage change in demonstration quality for each feedback condition: positive \textit{Improvement} reveals that the human is demonstrating the task more accurately.}

\begin{table*}[b]

    \caption{Questions on the Likert scale survey from Section~\ref{sec:vt2}. We grouped questions into five scales and examined their reliability using Cronbach's $\alpha$. Questions explored whether participants thought the robot's feedback was \textit{easy} to interpret, if they could \textit{focus} on teaching, how \textit{distinguishable} the robot's signals were, which methods were \textit{intuitive}, and their overall \textit{preferences}. For \textit{preference} we did not check for reliability since there was only a single item. We then performed a one-way repeated measures ANOVA on the grouped scores: here an $*$ denotes statistical significance.}
	\label{table:likert}
	\centering
		\begin{tabular}{lcccc}
			\hline Questionnaire Item & Reliability & $F(2,22)$ & p-value \bigstrut \\ \hline 
            \bigstrut[t]
            -- It was hard to figure out what the robot was trying to convey to me. & \multirow{2}{*}{$.75$} & \multirow{2}{*}{$1.699$} & \multirow{2}{*}{$.206$} \\  -- I could \textbf{easily} tell what the robot wanted. \bigstrut[b] \\ \hline  
            \bigstrut[t]
            -- I could \textbf{focus} on the robot's feedback without having to look up or move my hands. & \multirow{2}{*}{$.74$} & \multirow{2}{*}{$6.266$} & \multirow{2}{*}{$<.01^{*}$} \\ -- I had to physically go out of my way to get the robot's feedback. \bigstrut[b] \\ \hline  
            \bigstrut[t]
            -- It was easy to \textbf{distinguish} the different feedback signals. & \multirow{2}{*}{$.64$} & \multirow{2}{*}{$1.733$} & \multirow{2}{*}{$.215$} \\ -- I had to think carefully about what I was seeing / feeling to determine the signal. \bigstrut[b] \\ \hline  
            \bigstrut[t]
            -- The way the robot provided feedback seemed \textbf{intuitive} to me. & \multirow{2}{*}{$.86$} & \multirow{2}{*}{$.081$} & \multirow{2}{*}{$.923$} \\ -- I thought the robot's feedback was unintuitive and hard to understand. \bigstrut[b] \\ \hline  
            \bigstrut[t]
            -- Overall, I \textbf{prefer} this communication modality. & \multirow{1}{*}{$-$} & \multirow{1}{*}{$5.189$} & \multirow{1}{*}{$.191$} \bigstrut[b] \\ \hline   
		\end{tabular}
\vspace{-0.5em}
\end{table*}

\p{Dependent Measures -- Subjective} Participants responded to a $7$-point Likert scale survey after each feedback condition. Our survey was composed of four multi-item scales and one single-item scale (see Table~\ref{table:likert}). We asked participants how \textit{easy} it was to understand the robot's feedback, whether they could \textit{focus} on the task, how \textit{distinguishable} was the robot's feedback, if the feedback was \textit{intuitive} for this task, and to what extent they \textit{prefer} this condition as a communication modality. Finally, after participants had finished working with all the conditions they responded to a forced-choice comparison: ``Which method did you like the most?''

\p{Hypotheses}
We had two hypotheses for this user study:
\begin{displayquote}
    \textbf{H3.} \emph{Distributing multi-DoF haptic feedback along the robot arm (\textbf{Global}) will lead to improved demonstrations and lower teaching time.} 
\end{displayquote}
\begin{displayquote}
    \textbf{H4.} \emph{Participants will prefer distributed feedback (\textbf{Global}) as compared to localized feedback (\textbf{Local}).} 
\end{displayquote}

\begin{figure*}[t]
	\begin{center}
		\includegraphics[width=1.9\columnwidth]{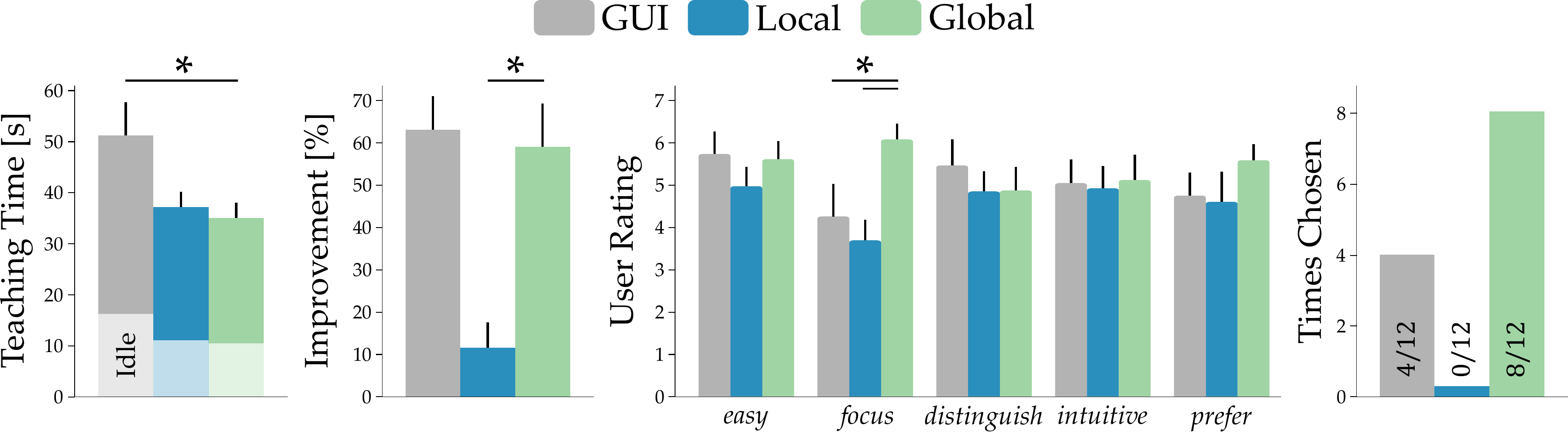}
		\caption{Objective and subjective results when communicating multi-dimensional robot feedback. We compared using a computer monitor (\textbf{GUI}), localizing wrapped haptic feedback to specific parts of the robot (\textbf{Local}), and distributing $3$-DoF feedback along the arm (\textbf{Global}). Error bars show standard error of the mean and an $*$ indicates statistically significant comparisons. (Left) Participants spent less time teaching the robot with \textbf{Global} as compared to \textbf{GUI}: shaded regions show the amount of time where participants stopped moving the robot to think about their next actions. The human's demonstrations improved more with \textbf{Global} feedback as compared to \textbf{Local} feedback. (Middle) Participants perceived the multi-DoF wrapped haptic display as similar to the alternatives, but indicated that \textbf{Global} enabled them to focus on teaching the robot. (Right) At the end of the experiment users were asked to choose their favorite method. Of the $12$ total participants, $8$ selected \textbf{Global}, $4$ selected \textbf{GUI}, and none selected \textbf{Local}.}
		\label{fig:results_study4}
	\end{center}
	\vspace{-1.5em}
\end{figure*}

\p{Results -- Objective} The results from this user study are summarized in \fig{results_study4}. To get a sense of the users' experience, we also show participant demonstrations in \fig{example_study4}.

Let us start our analysis by looking at the objective results. Using a one-way repeated measures ANOVA, we determined that feedback type had a significant effect on \textit{Teaching Time} ($F(2,22)=3.423$, $p<.05$). Post hoc tests revealed that participants spent less time demonstrating the task with \textbf{Global} than with \textbf{GUI} ($p<.05$), while the differences between \textbf{Global} and \textbf{Local} were not significant ($p=.675$). To explain these results we measured the amount of idle time during the demonstration. We found that with \textbf{GUI} users needed to stop, look at the monitor, and think about their next action: shifting attention back-and-forth between the monitor and the welding task contributed to the increased \textit{Teaching Time}.

So with \textbf{Global}, participants taught the robot more quickly --- but did they provide accurate, informative demonstrations? Remember that to measure \textit{Improvement} we first collected a demonstration without feedback, and then compared that initial demonstration to the user's behavior under each condition. The type of robot feedback had a significant effect on \textit{Improvement} ($F(2,22)=12.707$, $p<.001$). With both \textbf{GUI} and \textbf{Global} the participants made similar improvements to their teaching ($p=.769$). However, \textit{Improvement} was significantly lower for \textbf{Local} as compared to \textbf{Global} ($p<.01$). 
When participants received \textbf{Local} feedback they frequently had to change their grasp and move their hands across the three haptic displays; by contrast, in \textbf{GUI} and \textbf{Global} the participants could maintain a fixed grasp (\fig{example_study4}). 
Overall, our objective results support \textbf{H3}. \textbf{Global} enabled users to teach robots more seamlessly than \textbf{GUI} and more accurately than \textbf{Local}.

\p{Results -- Subjective} Table~\ref{table:likert} and \fig{results_study4} outline the results of our Likert scale survey and forced-choice comparison. We first checked the reliability of our four multi-item scales: \textit{easy}, \textit{focus}, and \textit{intuitive} were reliable (Cronbach's $\alpha > 0.7$) but \textit{distinguish} was not. We then grouped each scale into a combined score and performed a one-way repeated measures ANOVA on the result. Note that we did not check for reliability in \textit{prefer} because we only had one item (i.e., one question) on this scale.

We found that participants perceived \textbf{GUI}, \textbf{Local}, and \textbf{Global} to be similar along several axes. For instance, users did not think that any of the feedback types were more distinguishable ($p=.215$) or intuitive ($p=.923$) than the others. However, users reported that they were better able to \textit{focus} on the task with \textbf{Global} than with \textbf{GUI} ($p<.05$) or with \textbf{Local} ($p<.001$). After the experiment was finished we asked users to select their favorite feedback type: eight of the twelve participants chose \textbf{Global}, and the remaining four selected \textbf{GUI}. These subjective results support \textbf{H4}. \textcolor{black}{Given the results in Section VI, we were particularly interested to find that participants preferred \textbf{Global} feedback over \textbf{Local} feedback --- this suggests that the convenience of having all three signal available at each contact point along the arm in the \textbf{Global} feedback condition outweighed the slight decrease in accuracy.} One participant mentioned that ``\textit{I liked \textbf{Local} the least, since it requires repositioning hands to get feedback.}''

%% file: conclusion.tex
\section{Conclusion}

In this paper we presented a novel approach for communicating a robot's internal state during physical interaction. Specifically, we introduced a class of soft, wrapped haptic displays that are mounted on the robot arm at the point of contact between the human and robot; these displays provide real-time feedback as the robot learns from human demonstrations. 

We first designed wrapped pneumatic devices using flexible pouches that render one or more pressure signals (Section~\ref{Haptic Display and Design}). We performed psychophysics and robotics experiments with (a) $1$-DoF displays and (b) N-DoF displays. With the $1$-DoF display, humans could accurately distinguish between different pressures rendered by the wrapped haptic display (Section~\ref{sec:p1}). This feedback enabled participants to kinethetically teach robot arms more rapidly and effectively as compared to the alternatives (Section~\ref{sec:vt1}).

We next explored N-DoF haptic displays to communicate more detailed feedback. We compared two approaches: localizing separate $1$-DoF haptic displays to different regions of the robot arm, or distributing identical $3$-DoF displays along the entire arm. From a psychophysics perspective, localized feedback resulted in more accurate communication but at slower speeds, i.e. larger spatially distribution of signals increased accuracy, but required participants to move their hands to perceive each region, and recognize the signal (Section~\ref{sec:p2}). We applied both types of haptic displays to a robot learning task. Here we found that distributed $3$-DoF signals were preferable to localized $1$-DoF signals in terms of teaching time, demonstration improvement, and subjective responses (Section~\ref{sec:vt2}). \textcolor{black}{We also note that the results of our user studies were consistent across two different industry-standard robot arms, suggesting that approach is not tied to one specific arm type or geometry.} Overall, using multi-DoF haptic displays to concentrate signals into a smaller space resulted in more seamless communication and teaching.

\textcolor{black}{Future work will focus on further increasing the complexity of signals that the soft wrapped haptic displays can render. The stacking of pneumatic pouches developed by Do et al. \cite{do2021macro} may allow better spatial resolution by eliciting separate force and contact area signals. The integration of sensing technology to the concept of localized pressure distribution may also allow us to break down local pressure measurements (i.e. local user forces) and map them into desired robot motions.}
